\documentclass[journal]{IEEEtran}
\usepackage{hyperref}
\hypersetup{colorlinks,linkcolor={blue},citecolor={blue},urlcolor={green}}
\usepackage{amsmath,amsfonts}
\usepackage{amssymb}
\usepackage{amsthm}
\usepackage{algorithmic}
\usepackage{algorithm}
\usepackage{array}
\usepackage[caption=false,font=normalsize,labelfont=sf,textfont=sf]{subfig}
\usepackage{textcomp}
\usepackage{stfloats}
\usepackage{url}
\usepackage{verbatim}
\usepackage{graphicx}
\usepackage{cite}
\usepackage{mathabx}
\usepackage{comment}
\usepackage{booktabs}
\usepackage{multirow}
\usepackage{xcolor}
\usepackage{orcidlink}
\usepackage{pifont}

\usepackage{hyperref}
\hypersetup{colorlinks,linkcolor={blue},citecolor={blue},urlcolor={green}}
\begin{document}

\title{ParaBlock++: Communication-Computation Parallel Blockwise Hierarchical Federated Learning}

\author{Kai~Wang$^{\orcidlink{0000-0002-6170-4744}}$ and Chee~Wei~Tan$^{ \orcidlink{0000-0002-6624-9752}}$

\thanks{This work was supported in part by the Singapore Ministry of Education under Grant MOE-T2EP20224-0009. (Corresponding author: Chee Wei Tan.)}
\thanks{Kai Wang and Associate Professor Chee Wei Tan are with the College of Computing and Data Science, Nanyang Technological University, Singapore 639798 (e-mail: wang1899@e.ntu.edu.sg; cheewei.tan@ntu.edu.sg).}
\thanks{This work has been submitted to the IEEE TMC for possible publication. Copyright may be transferred without notice, after which this version may no longer be accessible.}
}



\maketitle

\begin{abstract}
In the emerging 6G paradigm, where AI-driven services operate over decentralized and resource-constrained infrastructures, the convergence of distributed learning algorithms remains a significant challenge. This paper introduces $\mathrm{ParaBlock}$++, a low Earth orbit (LEO) satellite-assisted hierarchical federated learning (HFL) framework, enabling resource-limited terrestrial clients to collaboratively engage in federated learning (FL) via ground-satellite links (GSLs) supported by LEO satellite megaconstellations. Within this architecture, LEO satellites serve multiple roles: as edge servers aggregating and broadcasting models, as clients accelerating training, and as high-throughput relays managing inter-satellite routing through inter-satellite links (ISLs). Departing from conventional synchronous Jacobian-type updates, $\mathrm{ParaBlock}$++ employs asynchronous BCD-type updates to enhance convergence speed. We provide theoretical convergence guarantees for convex, strongly convex, and non-convex objectives, and formulate optimization problems to determine optimal cluster resource allocation and model partitioning for minimizing global training latency. Experimental evaluations demonstrate that $\mathrm{ParaBlock}$++ achieves faster initial convergence and comparable final performance for deep neural networks (DNNs) such as ResNet-18, offering promising directions for federated learning across geographically remote areas enabled by 6G technologies, but much less global rounds and communication cost.
\end{abstract}


\section{Introduction}

\IEEEPARstart{T}{he} sixth-generation (6G) wireless networks are envisioned to push forward 5G capabilities by offering ultra-high bitrates, extremely low latency, pervasive artificial intelligence (AI), native support for ubiquitous connectivity across terrestrial, aerial, and satellite platforms, etc. These ambitions have been driven by emerging applications such as immersive extended reality (XR) \cite{xr}, digital twins \cite{digitaltwin}, and massive-scale internet-of-things (IoT) \cite{iot}. In particular, the increasing demand for intelligent applications at the wireless network edge has motivated the development of distributed learning paradigms capable of operating under constrained and heterogeneous environments. In the 6G landscape \cite{6g}, where network edge plays an important role as data containers and model trainers, designing distributed algorithms that are both communication-efficient and convergence-guaranteed present a unique challenge.

Federated learning (FL) \cite{fl1,fl2,fl3,fl4,fl5,fl6}, a decentralized machine learning framework in which edge devices collaboratively train a global model without sharing clients' raw data, has emerged as a promising technique and has received a lot of attention. Leveraging the computation capabilities of distributed equipment, the fusion of FL and 6G represents a critical enabler for truly intelligent, autonomous, and privacy-preserving wireless systems. FL is expected to serve not just as a service over 6G, but as a foundational design principle that shapes the evolution of future network architectures.

The emerging 6G techniques enable us to consider FL over remote geographical areas that are traditionally underserviced with terrestrial communication infrastructure and could be serviced by satellites, forming non-terrestrial networks (NTNs). In particular, we consider the typical situation in which clients in remote areas are grouped into clusters that are serviced by low earth orbit (LEO) satellites. We aim to involve all clients since their individual datasets could be highly non-i.i.d. across different clusters; meanwhile, the clients might have disproportionate computing capabilities that implies a potential split of training task over all clients accordingly.

Distributed clients and clusters require communication of model parameters, which suffers from scalability issues with single-server (i.e., centralized) FL. To mitigate this issue, we propose to use the hierarchical federated learning (HFL) architecture \cite{hfl1, hfl2, hfl3, hfl4, hfl5}, introducing edge servers as middle level aggregators in the hierarchy and reduces direct communication with the cloud server. With HFL, clients communicate with edge servers, which in turn communicate with the cloud server. Although conventional FL has demonstrated success across various applications, it is well-known that federated block coordinate descent (BCD) schemes \cite{fedbcd1, fedbcd2, fedbcd3, fedbcd4, fedbcd5} empirically achieve faster convergence. We integrate BCD-type updates into the HFL framework, allowing a cluster of clients to update a subset of model parameters and pass the updated parameters to subsequent clusters. This scheme is communication-efficient since a significant portion of submodel propagation could be completed in parallel with training, which is desirable for clients with limited communication bandwidth for submodels requiring long-distance transmission.

Motivated by the promising 6G techniques and inspired by a communication-efficient algorithm $\mathrm{ParaBlock}$ \cite{fedbcd5}, we propose $\mathrm{ParaBlock}$++, a BCD-type HFL framework that provides a simple yet effective way to address the aforementioned challenges, by interleaving computation with communication on resource-constrained clients. Compared with the preliminary version \cite{hafl}, we have presented a thorough analysis on communication efficiency, incorporated extensive evaluations, and provided detailed proof of convergence in this paper. The key contributions of this paper are summarized as follows:
\begin{itemize}
    \item We propose $\mathrm{ParaBlock}$++, a novel HFL framework to address communication latency during the federated BCD updates. $\mathrm{ParaBlock}$++ avoids the traditional sequential upload-aggregation-download procedure by parallelizing a significant amount of communication with training, effectively reducing the overall latency. This BCD-type method sequentially updates designated submodels in a randomly shuffled fashion, accelerating the training and convergence of models.
    \item We leverage LEO satellites in multiple orbit planes to assist geographically distributed clients in remote areas to participate in FL with their local datasets. The LEO satellites are versatile to perform submodel training with offloaded data like an ordinary client, orchestrate intra-cluster submodel aggregation via ground-satellite links (GSLs), and transmit trained submodels to remote LEO satellites via inter-satellite links (ISLs).
    \item We formulate a network latency minimization problem with multiple MILPs and solve for the system parameters, i.e., satellite CPU frequency, amount of offloaded data from clients to satellites, client bandwidth allocation, and submodel partition. We convert the problem equivalently into multiple independent MILPs and solve it using BCD, to avoid the generally NP-hard MILP formulation.
    \item We rigorously prove convergence guarantees of model parameters under convex, strongly convex, and non-convex objectives with bounded inter-cluster and intra-cluster gradient variance. We demonstrate that, under standard assumptions of e.g. Lipschitz continuity and bounded variance, $\mathrm{ParaBlock}$++ admits a convergence rate of $\mathit{O}(1/\sqrt{T})$ for convex objectives, $\mathit{O}(1/T)$ for strongly convex objectives, where $T$ is the number of global updates. $\mathrm{ParaBlock}$++ also ensures a vanishing gradient norm $\mathit{o}(1)$ for non-convex objectives.
    \item We conduct extensive experiments with various classification models to demonstrate the superior performance of $\mathrm{ParaBlock}$++. Specifically, we evaluate the classification accuracy on FashionMNIST, a modern and challenging image dataset. Our results show that $\mathrm{ParaBlock}$++  substantially reduces both the number of global training rounds and the total wall-clock time (computation and communication) required to attain a given test accuracy, highlighting its ability to enhance efficiency by leveraging potential computation-communication parallelism.
\end{itemize}

The remainder of the paper is organized as follows. Section \ref{sec:relatedwork} introduces related work on BCD-type updates, hierarchical federated learning, and satellite-assisted federated learning. Section \ref{sec:model} provides an overview of $\mathrm{ParaBlock}$++, our proposed parallel training and transmission model. Section \ref{sec:system} presents system description and problem formulation on the communication aspect, which is solved in Section \ref{sec:iterative}. Section \ref{sec:hfl} introduces the HFL aspect of the $\mathrm{ParaBlock}$++ model, whose convergence analysis is provided in Section \ref{sec:convergence}. Section \ref{sec:results} presents experiment results and analysis. Section \ref{sec:conclusion} concludes the paper with discussion and future work.

\section{Related Work}\label{sec:relatedwork}

\subsection{Block Coordinate Descent Update for FL}

Compared to conventional Jacobian-type gradient descent, BCD empirically accelerates convergence because it makes more informed progress per iteration through Gauss-Seidel's rule based on freshly updated parameters. The principle of BCD dates back to early machine learning literature with a lineage that connects to classical numerical analysis \cite{bcd1, bcd2, bcd3}, and has seen a significant resurgence in modern optimization \cite{bsg, bcd4}. Combined with FL, federated BCD has demonstrated remarkable performance in solving large-scale optimization tasks and has attracted research attention \cite{fedbcd1, fedbcd2, fedbcd3, fedbcd4}. Recent advances in BCD-type FL have been leveraged especially for efficient fine-tuning of large language models (LLMs) \cite{fedbcdllm1, fedbcdllm2, fedbcd4}.

\subsection{Hierarchical Federated Learning}

HFL has been proposed to address scalability and communication challenges by introducing multiple layers of aggregation, typically involving clients, edge servers, and central servers. Early works such as HierFAVG \cite{hierfavg} demonstrated that hierarchical aggregation can significantly reduce communication overhead and improve convergence speed, particularly in environments with limited client-server bandwidth. Several extended works account for heterogeneity across clients and network tiers. For example, HierFAVG \cite{hierfavg} was proposed to better balance local and global updates, while other works \cite{hfl6} analyzed convergence bounds under non-i.i.d. data distributions. Personalized HFL \cite{hfl7} was proposed to improve model accuracy across diverse client populations using local fine-tuning and cross-layer aggregation. Recent works have explored communication-efficient techniques for HFL, such as \cite{hfl8, hfl9}. In the era of LLMs, potential latencies due to communication of massive-scale model parameters gradually become the bottleneck of efficient training and deployment of LLMs and thus deserves deliberation.

\subsection{Satellite-Assisted Federated Learning}

Satellite-assisted federated learning (SAFL) has emerged as a promising solution to extend FL to remote and globally distributed users. Using satellite networks, particularly LEO satellite megaconstellations, SAFL can overcome limitations of terrestrial infrastructures, providing global coverage and improved accessibility for traditionally underserviced areas. Recent works have proposed various hierarchical architectures, in which satellites serve as intermediate aggregators, reducing communication overhead and addressing connectivity constraints \cite{satellite5}. However, SAFL introduces new challenges including intermittent links, data heterogeneity, and security risks. Furthermore, the use of ISLs has been explored to enhance model dissemination and system robustness. However, prior works \cite{satellite6,satellite7,satellite8,satellite9} on SAFL that associate terrestrial clients in remote geographical areas are limited. Our setting with client-local datasets offloaded to satellites is significantly different from the settings in which each satellite directly collects its local data. 

\section{Concurrent Training and Transmission Model}\label{sec:model}

The key benefit of our proposed $\mathrm{ParaBlock}$++ model relies on an advanced, 6G-enabled NTN to aggregate, broadcast, and propagate asynchronously updated submodels. In this section, we will explain the modules of this framework.

\subsection{The Satellite Communication System}

The hierarchical learning and communication system consists of cloud servers $\mathcal{K}$, edge servers $\mathcal{I}$, client devices $\mathcal{J}$, and clusters $\{\mathcal{C}_i\}_{i\in\mathcal{I}}$. The clusters are geographically dispersed across remote areas, with each cluster consisting of spatially co-located clients. Each client $j \in \mathcal{J}$ is associated with exactly one cluster $\mathcal{C}_i$, such that $\mathcal{J} = \cup_i \mathcal{C}_i$, $\mathcal{C}_i \cap \mathcal{C}_{i'} = \varnothing$ for all $i \neq i'$. Furthermore, each cluster $\mathcal{C}_i$ is associated with a corresponding edge server $i \in \mathcal{I}$, although an edge server may concurrently serve multiple clusters.

Clients are resource-constrained terrestrial user equipment (e.g., 6G-enabled smartphones) that can participate in FL only indirectly, e.g., via LEO satellite links. Edge servers (e.g., LEO satellites) are responsible for orchestrating training and communication within their respective clusters, while cloud servers (e.g., a set of logically centralized GEO satellites) oversee global management and model aggregation across the spatially distributed clusters. Given the widespread deployment and rapidly declining cost of satellite constellations in the 6G era, it is ensured that any geographic location is continuously covered by at least one LEO satellite.

We adopt the LEO megaconstellation with LEO satellites in $P$ orbit planes \cite{satellite}, where each orbit consists of $Q$ co-orbital LEO satellites that travel around the Earth on the same orbit in sequence. Since LEO satellites are not geostationary, we assume that the entire Earth surface is covered by static logic locations \cite{satellite2}. To ensure continuity of service at these locations, the satellite currently serving the cluster must hand over its services to the next co-orbital satellite. The GEO satellites function as cloud servers, which exercise management functionalities besides global model aggregation, such as wireless routing strategies among all LEO satellites and LEO satellites status information bookkeeping.

Related work \cite{satellite2, satellite3, satellite4} proposed to model the LEO satellite megaconstellation as fixed topology referred to as topology snapshot within a short time slot $T$. In a given snapshot, the GEO satellite can arrange the routing scheme using ISL. Given the altitude $h_n$ of orbit $n$ and Earth radius $r_e$, $h_n^*=h_n+r_e$ is the distance from orbit $n$ to Earth center, and the position satellite $i$ in orbit $n$ in geocentric coordinates is given by \cite{satellite2}
\begin{align*}
    x_i=&\ h_n^*\cos\alpha_i\cos\gamma_n-h_n^*\sin\alpha_i\cos\kappa_n\sin\gamma_n,\\
    y_i=&\ h_n^*\cos\alpha_i\sin\gamma_n+h_n^*\sin\alpha_i\cos\kappa_n\sin\gamma_n,\\
    z_i=&\ h_n^*\sin\alpha_i\cos\kappa_n,
\end{align*}
where $\alpha_i$, $\gamma_n$, and $\kappa_n$ denote the right ascension of ascending node of satellite $i$, the longitude angle, and the inclination of orbit $n$, respectively. The distance $d_{i,i'}$ between satellites $i$ and $i'$ is given by
\begin{align*}
    d_{i,i'}=\sqrt{(x_i-x_{i'})^2+(y_i-y_{i'})^2+(z_i-z_{i'})^2},
\end{align*}
and the communication radius of satellite $i$ in orbit $n$ is
\begin{align*}
    d^\text{comm}_i=2\sqrt{(h_n^*)^2-r_e^2}.
\end{align*}
The inter-orbit satellites $i$ and $i'$ can communicate via ISL if and only if they are both within each other's communication radius, i.e., $d_{i,i'}\leq\min(d^\text{comm}_i,d^\text{comm}_{i'})$ \cite{satellite2}. Meanwhile, we also make assumptions that intra-orbit ISLs are only available between adjacent satellites in the same orbit \cite{satellite2}. With the set of available ISLs, we are able to construct the communication graph for routing the submodels.

\subsection{The Satellite Handover and Communication Model}

LEO satellites are versatile and serve multiple purposes, including acting as clients for submodel training, edge servers for intra-cluster model aggregation, or relays for inter-cluster model transmission. LEO satellites are non-geostationary that provide coverage to a cluster only for coverage time $T$, after which the connection and task is handed over to a neighboring satellite in time $\tau^{\text{trans}}_{S,i}$. We adopt the recently proposed satellite handover model \cite{gts}
\begin{align}\label{eq:handover}
    T^\text{eff}_{S,i}=T^\text{raw}_{S,i}+(N_i+1)\cdot\tau^\text{trans}_{S,i},\ N_i=\Big\lfloor\frac{T^\text{raw}_{S,i}}{T-\tau^\text{trans}_{S,i}}\Big\rfloor,
\end{align}
where $T^\text{raw}_{S,i}$ is the raw task time without handover and $T^\text{eff}_{S,i}$ is the effective task time to completion. The rationale is that, satellite $i$ must reserve time $\tau^\text{trans}_{S,i}$ at the end of covering slot $T$ to transmit its model and dataset to an adjacent satellite, which could continue with the current task thereafter. Satellite $i$ can communicate to satellite $j$ via ISL with rate given by
\begin{align}\label{eq:srate}
    R^\text{ISL}_{i,j}=B_{S,i}\log_2\Big(1+\frac{P_{S,i}|h_{i,j}|^2}{B_{S,i}N_0}\Big),
\end{align}
where $B_{S,i}$ is the bandwidth for satellite $i$, $N_0$ is the white noise power spectral density, $P_{S,i}$ is satellite transmit power, and $h_{i,j}$ is the corresponding channel coefficient. For ISLs where free-space LoS channels dominate, we model $h_{i,j} = \sqrt{\beta_{i,j}}$ and model $\beta_{i,j}$ as
\begin{align}\label{eq:lossmodel}
    \beta_{i,j}=\beta_0-10\cdot\alpha^{\text{ISL}}\cdot\log_{10}\Big(\frac{d_{i,j}}{d_0}\Big),
\end{align}
where $\beta_0 = -30$dB denotes the large-scale pathloss coefficient at a reference distance of $d_0 = 1$m, $d_{i,j}$ denotes the distance between satellites $i$ and $j$, and $\alpha^{\text{ISL}}\!=\!2.00$ denotes the pathloss exponent for free-space ISLs.

For GSL channels that exist in urban areas between satellite and client, we incorporate both large-scale and small-scale fading effects in $h_{i,j}$, given by $h_{i,j}=\sqrt{\beta_{i,j}}u_{i,j}$, where $\beta_{i,j}$ is the large-scale pathloss coefficient and $u_{i,j}\sim\mathcal{CN}(0,1)$ captures Rayleigh fading. For urban areas, we replace $\alpha^{\text{ISL}}$ in Eq. \eqref{eq:lossmodel} with $\alpha^{\text{GSL}}=3.50$, denoting the pathloss exponent for downlink and uplink GSL channels in urban areas, resp.

The energy consumption for transmission via ISL is
\begin{align}\label{eq:senergy}
    E^\text{ISL}_{i,j} = P_{S,i} \cdot \tau^\text{ISL}_{i,j}.
\end{align}
Similar formulas for transmission rate and energy hold for GSL channels. For relaying, satellite $i$ is unaffected by handover, since it need not store the relayed data, and the relay target satellite $i'$ must belong to a different cluster $i'\neq i$, ensuring that the relay and any potential handover do not share the same communication channel. Besides, we assume that satellites can perform computation while transmitting data (except during handover).

\subsection{Communication System Optimization Objective}

Our objective is to optimize the communication system with respect to the total time required for the entire training and communication process. This stands in contrast to prior work (e.g., \cite{gts}) that focused on optimizing specific FL performance lower bounds. Given that such bounds are often loose with substantial hidden factors that do not accurately align with real-world performance, we argue that optimizing the communication system based on communication-related objectives (e.g., total delay or latency) is more practical and sound. The system design is optimized with the following sets of hyperparameters.

\textit{1) Model Partition}. The $\mathrm{ParaBlock}$++ system employs distributed clusters for collaborative training with model partitioning, improving training efficiency with better parallelism on multiple devices. Model partitioning also provides an essential technique for low-resource client devices to participate in FL. Our system splits the large model into $s$ mutually disjoint submodels, and we require $s=|\{\mathcal{C}_i\}_{i\in\mathcal{I}}|$ for ease of implementation, such that there exists a one-to-one mapping between clusters and assignment of submodels. We aim to optimize the partition of model into sizes $\{|S_i|\}_{i=1}^s$ and each cluster $i$ is responsible for training submodel $S_i$. However, all clients must have a local version of the full model to train its assigned submodel.

\textit{2) Data Availability}. Each client $j$ has local dataset $\mathcal{D}_j\!=\!\mathcal{D}_j^c\bigcup\mathcal{D}_j^s$ with $\mathcal{D}_j^c\bigcap\mathcal{D}_j^s\!=\!\varnothing$, where $\mathcal{D}_j^c$ contains privacy-sensitive data that only allow local access and $\mathcal{D}_j^s$ contains non-sensitive data that could be offloaded to satellites. Satellites have no local data of their own except those offloaded from same-cluster clients. Assuming non-i.i.d. datasets across clusters and clients, our model essentially reflects the real-world features of decentralized, user-owned datasets. For each cluster $i$, we attempt to determine the set of optimal ratios $\{\alpha_j\}_{j\in\mathcal{C}_i}$ of offloaded data.

\textit{3) Bandwidth}. Each cluster $i$ has bandwidth $B_i$ to communicate with LEO satellites that is allocatable to all clients $j\!\in\!\mathcal{C}_i$. For each cluster $i$, we aim to optimize the set $\{B_{C,j}\}_{j\in\mathcal{C}_i}$.

\textit{4) Frequency}. Each satellite $i$ has a tunable CPU frequency $f_{S,i}$ that affects training time and energy consumption. For each cluster $i$, we aim to optimize $f_{S,i}$.

\textit{5) Routing Schemes}. The proposed communication system necessitates long-range inter-cluster submodel transmission and propagation across remote geographical areas at the end of each global round. By modeling the LEO satellite constellation with ISLs as a directed graph, we further aim to optimize routing to minimize latency.

\section{System Description and Problem Formulation}\label{sec:system}

In this section, we describe each stage of the system that occurs within a single global round. By convention, we use the 3-tuple superscript $(t, e, h)$ to denote variables of interest at global round $t$, cluster round $e$, and local round $h$, resp., and use symbols with base $T$ such as $T^{\text{cluster}}_{S,i}$ to denote longer duration and those with base $\tau$ such as $\tau^{\text{trans}}_{S,i}$ to denote shorter duration.

\subsection{Preprocessing: Data and Model Offloading to Satellites}

Before global round $t=0$ starts, each client $j\!\in\!\mathcal{C}_i$ offloads a subset of non-sensitive data $\mathcal{D}_j^c\!\subset\!\mathcal{D}_j$ to the satellite $i$. Define $\alpha_j$ as the portion of data offloaded to satellite and define $\alpha_j^{\max}=|\mathcal{D}_j^c|/|\mathcal{D}_j|$ as the maximum portion of data offloaded to satellite, i.e., non-sensitive data. The cloud server allocates submodel $S_i$ to cluster $i$, whose edge server (i.e., satellite $i$) ensures that all clients $j\!\in\!\mathcal{C}_i$ are synchronized with the same initialized submodel parameters for $S_i$. In the model, these steps are executed at all participants in a one-shot manner throughout the entire course, prior to the start of global round $t=0$.

\subsection{Cluster Round: Satellite Side Update}

We decide mini-batch sizes that scale proportionally to the local dataset size, enabling effective utilization of all on-device data under a fixed number of training rounds across datasets of varying sizes. For simplicity, we express the raw task time for $H$ rounds of submodel updates at satellite $i$ as
\begin{align}\label{eq:T_clusterraw_Si}
    T_{S,i}^\text{cluster,raw}=\frac{Hm_{S,i}\gamma_{S,i}\sum_{j\in\mathcal{C}_i}\alpha_j|\mathcal{D}_j|}{f_{S,i}},
\end{align}
where $f_{S,i}$ is the CPU frequency of satellite $i$, $m_{S,i}$ is the number of CPU cycles per sample, $\gamma_{S,i}$ is the portion of samples a mini-batch contains, and the handover time is
\begin{align}\label{eq:tau_trans_Si}
    \tau^{\text{trans}}_{S,i}=\frac{|S|+q\cdot\sum_{j\in\mathcal{C}_i}\alpha_j|\mathcal{D}_j|}{R^{\text{ISL}}},
\end{align}
where $|S|$ is the size of the entire model $S$ (always needed to train any submodel $S_i$), $q$ is the size of each data sample, and $R^{\text{ISL}}$ is the high-throughput ISL rate between nearby satellites that is assumed to be constant. The satellite handover gives the satellite side latency as
\begin{align}\label{eq:T_cluster_Si}
    T_{S,i}^\text{cluster}=T_{S,i}^\text{cluster,raw}+\Big(\Big\lfloor \frac{T_{S,i}^\text{cluster,raw}}{T-\tau^\text{trans}_{S,i}}\Big\rfloor+1\Big)\cdot{\tau^\text{trans}_{S,i}}.
\end{align}
The energy consumption per cluster round at satellite $i$ is the sum of the energy used for CPU and transmission
\begin{align}\label{eq:E_cluster_Si}
    E^\text{cluster}_{S,i}=\kappa T_{S,i}^\text{cluster,raw} f_{S,i}^3+P_{S,i}\cdot \Big(\Big\lfloor\frac{T_{S,i}^\text{cluster,raw}}{T-\tau^\text{trans}_{S,i}}\Big\rfloor+1\Big)\cdot{\tau^\text{trans}_{S,i}},
\end{align}
where the terms denote CPU computation and transmission energy consumptions, respectively. The energy constraint requires that the battery level of satellite $i$ should always remain above a given threshold. That is,
\begin{align*}
    \phi^{\text{initial}}_{S,i}-E^{\text{cluster}}_{S,i}+{P^{\text{sun}}_{S,i}}\cdot{T^{\text{cluster}}_{S,i}}\geq\phi^{\min}_{S,i},
\end{align*}
where $P^{\text{sun}}_{S,i}$ means the solar energy recharge rate at satellite $i$, and $\phi^{\text{initial}}_{S,i}$ and $\phi^{\min}_{S,i}$ denote the initial battery level charged and the minimum battery level to function at satellite $i$, resp. However, monitoring battery level is difficult because of frequent handovers and the presence of external solar energy recharge. We impose a more stringent constraint that the energy consumption of satellite $i$ should not exceed the energy recharge from the sun during a cluster round. That is,
\begin{align}\label{eq:E_cluster_Si_constraint}
    E^{\text{cluster}}_{S,i}\leq{P^{\text{sun}}_{S,i}}\cdot{T^{\text{cluster}}_{S,i}},
\end{align}
which corresponds to setting $\phi^{\text{initial}}_{S,i}=\phi^{\min}_{S,i}$.

\subsection{Cluster Round: Client Side Update and Upload}

The computation time $T_{C,j}^\text{cluster}$ to finish $H$ local updates at client $j$ is written as
\begin{align}\label{eq:T_cluster_Cj}
    T_{C,j}^\text{cluster}=\frac{Hm_{C,j}\gamma_{C,j}(1-\alpha_j)|\mathcal{D}_j|}{f_{C,j}},
\end{align}
where $f_{C,j}$ is the CPU frequency at client $j$, $m_{C,j}$ is the number of CPU cycles for client $j$ to process one sample, and $\gamma_{C,j}$ is the portion of samples a mini-batch contains. The corresponding energy consumption is given by
\begin{align}\label{eq:E_cluster_Cj}
    E_{C,j}^\text{cluster}=\kappa{H}m_{C,j}\gamma_{C,j}(1-\alpha_j)|\mathcal{D}_j|f_{C,j}^2.
\end{align}
The communication time for client $j$ to upload the submodel to satellite $i$ is
\begin{align}\label{eq:tau_agg_Cj}
    \tau_{C,j}^\text{agg}=\frac{|S_i|}{R^{CS}_{j,i}},\ {R^{CS}_{j,i}}=B_j\log_2\Big(1+\frac{P_{C,j}|h_{j,i}|^2}{B_jN_0}\Big).
\end{align}
The corresponding energy consumption for submodel transmission at client $j$ becomes
\begin{align}\label{eq:E_agg_Cj}
    E^\text{agg}_{C,j}=P_{C,j}\cdot\tau^\text{agg}_{C,j}.
\end{align}
The analysis of client-side updates and uploads alone is more nuanced, and we adopt the client latency model \cite{gts} to capture this complexity. The duration $T^{\text{total}}_{i}$ measures the time elapsed from the start of the cluster round to the point when satellite $i$ has aggregated submodels from all clients, as computed by Eq. \eqref{eq:T_total_i} under three cases in Eq. \eqref{eq:T_total_cases}.

\begin{figure*}[!t]
\vspace{-20pt}
\begin{align}\label{eq:T_total_i}
T^\text{total}_i = 
\begin{cases}
T \cdot N_i + \max_{j \in \mathcal{C}_i} \big\{ \tau^{\text{agg}}_{C,j} \big\}, & \text{ (case $a$)$\qquad\qquad\qquad\qquad\qquad\qquad\qquad$} \\
\max_{j \in C_i} \big\{ \max \big( T \cdot \big\lfloor \dfrac{T^{\text{cluster}}_{C,j'}}{T} \big\rfloor, T^{\text{cluster}}_{C,j} \big) + \tau^{\text{agg}}_{C,j} \big\}, & \text{ (case $b$)$\qquad\qquad\qquad\qquad\qquad\qquad\qquad$} \\
T \cdot \big( \big\lfloor \dfrac{T^{\text{cluster}}_{C,j'}}{T} \big\rfloor + 1 \big) + \max_{j \in C_i} \big\{ \tau^{\text{agg}}_{C,j} \big\}, & \text{ (case $c$)$\qquad\qquad\qquad\qquad\qquad\qquad\qquad$}
\end{cases}
\qquad\\
\begin{cases}\label{eq:T_total_cases}
    \text{case $a$:} & T^{\text{cluster}}_{C,j'}\leq T \cdot N_i \\
    \text{case $b$:} & T^{\text{cluster}}_{C,j'}> T \cdot N_i\text{ and }\max_{j \in C_i} \big\{ \max \big( T \cdot \big\lfloor \dfrac{T^{\text{cluster}}_{C,j'}}{T} \big\rfloor, T^{\text{cluster}}_{C,j} \big)\! +\! \tau^{\text{agg}}_{C,j} \big\}\!\leq\! T \cdot \big( \big\lfloor \dfrac{T^{\text{cluster}}_{C,j'}}{T} \big\rfloor\! +\! 1 \big) \\
    \text{case $c$:} & T^{\text{cluster}}_{C,j'}> T \cdot N_i\text{ and }\max_{j \in C_i} \big\{ \max \big( T \cdot \big\lfloor \dfrac{T^{\text{cluster}}_{C,j'}}{T} \big\rfloor, T^{\text{cluster}}_{C,j} \big)\! +\! \tau^{\text{agg}}_{C,j} \big\}\!>\! T \cdot \big( \big\lfloor \dfrac{T^{\text{cluster}}_{C,j'}}{T} \big\rfloor\! +\! 1 \big) 
\end{cases}
\end{align}
\vspace{-20pt}
\end{figure*}
In Eqs. \eqref{eq:T_total_i}-\eqref{eq:T_total_cases}, $j'\!=\!\arg\max_j\{T^\text{cluster}_{C,j}\}$ and $N_i$ is defined for satellite $i$ in Eq. \eqref{eq:handover}. By selecting system parameters such that $\tau^{\text{agg}}_{C,j}\leq T$ for all $j$, the satellite handover mechanism ensures submodel aggregation could be completed within coverage slot $T$. case ($a$) corresponds to that the $(N_i+1)$-th satellite will have the aggregated submodel and clients could start uploading at time $T\cdot N_i$, case ($b$) corresponds that the $(\lfloor T^{\text{cluster}}_{C,j'}/T\rfloor+1)$-th satellite will have the aggregated submodel and clients could start uploading at time $T\cdot\lfloor T^{\text{cluster}}_{C,j'}/T\rfloor$, and case ($c$) corresponds to that the $(\lfloor T^{\text{cluster}}_{C,j'}/T\rfloor+2)$-th satellite will have the aggregated submodel and clients could start uploading at time $T\cdot(\lfloor T^{\text{cluster}}_{C,j'}/T\rfloor+1)$.

The latency per cluster round $T^\text{cluster}_i$ is the larger of satellite latency and client latency, plus synchronization delay $\tau^\text{sync}_i$.
\begin{align}\label{eq:total_delay}
    T^\text{cluster}_{i}=\max(T_{S,i}^\text{cluster},T^\text{total}_i)+\tau^\text{sync}_i,
\end{align}
where $\tau^\text{sync}_i$ is the additional latency to broadcast aggregated submodel to all clients in cluster $i$. The cluster rounds are repeated $E$ times on both satellite and client sides, after each of which a consensus submodel is aggregated at satellite $i$ and broadcast to its clients, requiring a total delay of $E\cdot T^\text{cluster}_{i}$.

\subsection{Global Round: Inter-Cluster Propagation}

To enable inter-cluster propagation with ISL, we build the satellite constellation topology based on availability of ISLs that connect remote LEO satellites. When direct ISLs are not possible, data may be relayed through other LEO satellites in a multi-hop route. In addition, we assume that the ISL bitrates may experience mild variations, while the outage probabilities due to fading remain negligible and satellites are considered interconnected unless LoS communication is obstructed by the Earth. To construct the directed graph $\mathcal{G}=(\mathcal{V},\mathcal{E})$ denoting the LEO satellite communication network, we use the Starlink LEO megaconstellation \cite{starlink} with settings $53.0\text{\textdegree}:1584/72/3/550$ km and the GPS GEO constellation \cite{gps} with settings $55.0\text{\textdegree}:24/6/4/20,200$ km. Given these satellite topologies, we can analyze the routing strategies on the resulting communication graph $\mathcal{G}$.

\begin{figure}[t]
  \centering
  \includegraphics[width=0.50\textwidth]{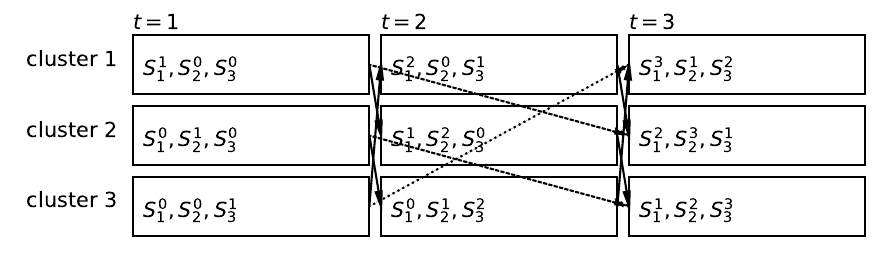}
  \vspace{-20pt}
  \caption{An illustration of the submodel transmission scheme. Solid lines denote the transmission stage and dotted lines denote the propagation stage. For example, cluster 2 needs $S^1_1$ to train $S^2_2$ at $t=2$ and hence $S^1_1$ must be received before $t=2$, corresponding to the transmission stage. Although cluster 3 also needs $S^1_1$ to train $S^3_3$ at $t=3$, it is time-insensitive and can be received during training at $t=2$, corresponding to the propagation stage.}
  \label{fig:transmissionscheme}
  \vspace{-10pt}
\end{figure}

To avoid confusion, we denote the submodel $S_i$ trained during global round $t$ as $S^t_i$. For all $i$, global round $t$ has two additional stages after completing the $E$ cluster rounds, as illustrated in Fig. \ref{fig:transmissionscheme}.

\noindent\textit{1) Transmission Stage}. In this stage, satellite $i$ transmits submodel $S_i^t$ to satellite $({i}\!\!\mod{s})+1$. This operation simply corresponds to solving a minimum hop count routing problem between a source–destination pair of LEO satellites within the constellation.

\noindent\textit{2) Propagation Stage}. In this stage, satellite $i$ propagates submodel $S_i^t$ to all other satellites $i'\!\neq\!(i\!\!\mod s)+1$ for use in subsequent iterations. This stage proceeds concurrently with the training of the submodel $S_{(i\!\!\mod s)+1}$ in global round $t+1$, during which satellite $i$ may continue transmitting the trained submodel $S^t_i$ to all other satellites. Fortunately, the time constraint is substantially more permissive, as the entire training process incurs no delay as long as the propagation completes before the conclusion of training submodel $S_{(i\!\!\mod s)+1}$, a condition readily satisfied given modern ISLs' ultra-high bitrates. We can formulate the problem as solving for the minimum delay over communication graph $\mathcal{G}\!=\!(\mathcal{V},\mathcal{E})$, subject to the requirement that satellite $i$ propagates $S^t_i$ to all other satellites\footnote{Strictly speaking, propagation to satellite $i\!\!\mod s+1$ may be omitted, as it has already received submodel $S^t_i$ during the transmission stage. However, this omission would render the problem NP-hard, reducing it to a minimum-cost directed Steiner tree (DST) problem.} without a prescribed order. As a practical heuristic, we model this problem as a minimum spanning arborescence (MSA) \cite{satellite,msa1,msa2} problem, the directed analogue of the minimum spanning tree (MST). Specifically, we construct an MSA rooted at each satellite $i$, using transmission delay as the link weight.


In the next section, we are committed to solving the latency minimization problem for each global round.

\section{Iterative Global Latency Minimization}\label{sec:iterative}

The latency of a global round $T^{\text{glob}}$ is given by
\begin{align}\label{eq:glob}
    T^{\text{glob}}=E \cdot \max_iT_i^{\text{cluster}}.
\end{align}
In minimizing $T^{\text{glob}}$, we use the assumption (e.g., in \cite{gts}) that sync delays $\tau^{\text{sync}}_i$ are short whose optimization contributes marginally to global latency minimization. This approximation is reasonable with high-throughput downlink GSLs since $T^{\text{glob}}$ is dominated by the first term of Eq. \eqref{eq:total_delay}.

The global latency minimization problem is equivalent to
\begin{align}
    \min_{\alpha, f_S, B_C, S}&\ \max_i\max\{T_{S,i}^{\text{cluster}},T_i^{\text{total}}\} \label{eq:Tglob} \\
    \text{s.t.:}&\ 0 \leq \alpha_j \leq \alpha_j^{\max},\ \forall j \in \mathcal{C}_i, \ \forall i \in \mathcal{I} \label{eq:Tglob1} \\
    &\ f_{S,i}^{\min} \leq f_{S,i} \leq f_{S,i}^{\max},\ \forall i \in \mathcal{I} \label{eq:Tglob2} \\
    &\ \textstyle\sum_{j \in \mathcal{C}_i} B_{C,j} \leq B_i,\ \forall j \in \mathcal{C}_i, \ \forall i \in \mathcal{I} \label{eq:Tglob3} \\
    &\ E_{C,j}^{\text{cluster}} + E_{C,j}^{\text{agg}} \leq \delta_j, \ \forall j \in \mathcal{C}_i,\ \forall i \in \mathcal{I} \label{eq:Tglob4} \\
    &\ E_{S,i}^{\text{cluster}} \leq P_i^{\text{sun}} \cdot T_{S,i}^{\text{cluster}}, \ \forall i \in \mathcal{I} \label{eq:Tglob5} \\
    &\ \textstyle\sum_{j \in \mathcal{C}_i} \alpha_j |\mathcal{D}_j| \leq A_i^{\max}, \ \forall i \in \mathcal{I}, \label{eq:Tglob6} \\
    &\ \textstyle\sum_{i\in\mathcal{I}}|S_i|=|S|. \label{eq:Tglob7}
\end{align}

We propose a BCD algorithm to address the non-convex problem in Eqs. \eqref{eq:Tglob}-\eqref{eq:Tglob7}, which iteratively optimizes over $\alpha=[\alpha_{j}]_{j\in\mathcal{C}_i,i\in\mathcal{I}}$, $f_S=[f_{S,i}]_{i\in\mathcal{I}}$, $B_C=[B_{C,j}]_{j\in\mathcal{C}_i,i\in\mathcal{I}}$, and $S=[S_i]_{i\in\mathcal{I}}$, respectively. The use of BCD is a prevalent approach in solving optimization problems that are non-convex across all variables yet block-wise convex with respect to individual variable subsets. Its primary rationale, however, lies not in accelerating convergence, but in enhancing mathematical tractability. Given an arbitrary $t$, we can check the feasibility of $\max_i\max\{T_{S,i}^{\text{cluster}},T_i^{\text{total}}\}\leq t$ with all the constraints in Eqs. \eqref{eq:Tglob1}-\eqref{eq:Tglob7}, which implicitly include two integer equations
\begin{align}\label{eq:integers}
    N_i=\Big\lfloor\frac{T_{S,i}^\text{raw}}{T-\tau_{S,i}^\text{trans}}\Big\rfloor, \
    M_i=\Big\lfloor\frac{T_{C,j'}^\text{cluster}}{T}\Big\rfloor
\end{align}
and disjoint cases formulated by Eq. \eqref{eq:T_total_i} and Eq. \eqref{eq:T_total_cases}. The non-linearity introduced by Eq. \eqref{eq:integers} is efficiently handled by MILP solvers, as the ranges of $N_i$ and $M_i$ remain very limited for practical choices of $T$ (typically on the order of hundreds of seconds). Likewise, the complications arising from Eq. \eqref{eq:T_total_i} and Eq. \eqref{eq:T_total_cases} are readily resolved by a straightforward lemma that reduces the number of cases for $T_i^\text{total}$.\newline

\noindent\textbf{Lemma} (Reduction to disjoint cases with convex constraints). To check the feasibility of $T^{\text{total}}_i\ \!{\leq}\ \!t$, we can sequentially check it under cases ($a$), ($b$), and ($c'$), resp., where case ($c'$) has all but the last constraint of case ($c$), i.e.,
\begin{align*}
    \max_{j\in\mathcal{C}_i}\big\{\max \big( T \cdot \big\lfloor \dfrac{T^{\text{cluster}}_{C,j'}}{T} \big\rfloor, T^{\text{cluster}}_{C,j} \big) {+} \tau^{\text{agg}}_{C,j}\big\}> T\cdot\big(\big\lfloor \dfrac{T^{\text{cluster}}_{C,j'}}{T} \big\rfloor {+} 1\big).
\end{align*}
The MILP reduces into a number of feasibility tests with convex constraints in the form of $\max\{\text{expr}_1,\text{expr}_2,\ldots\}\leq\text{expr}_3$.

\noindent\textit{Proof}: We consider case ($c$) only when neither case ($a$) nor case ($b$) holds. We show that if cases ($a$) and ($b$) do not satisfy Eq. \eqref{eq:Tglob}, then the last constraint of case ($c$) can be safely omitted. If the problem is infeasible under case ($c'$), then it must also be infeasible under case ($c$), as the latter imposes additional constraints. Conversely, if the problem is infeasible under case ($c$), then it is necessarily infeasible under case ($c'$). Otherwise, the situation reduces to case ($b$) in Eq. \eqref{eq:T_total_cases}, but the objective values under cases ($c$) and ($c'$) are both greater than or equal to that under case ($b$), implying that case ($b$) must satisfy Eq. \eqref{eq:Tglob}. This yields a contradiction. Consequently, it suffices to sequentially check feasibility for cases ($a$), ($b$), and ($c'$), respectively. \hfill\qedsymbol

\subsection{Optimize $\alpha$ given $f_S$, $B_C$, and $S$}

This subproblem corresponds to finding the lowest value of $t$ and corresponding $\boldsymbol{\alpha}$ for which problem \eqref{eq:optalpha} is feasible.
\begin{align}\label{eq:optalpha}
    \min_{\alpha}\ 1,\ \text{s.t.: }&\ \max_iT_i^\text{total}\leq t,\notag \\
    &\ \max_iT^{\text{cluster}}_{S,i}\leq t,\eqref{eq:Tglob1},\eqref{eq:Tglob4},\eqref{eq:Tglob5},\eqref{eq:Tglob6}
\end{align}
We can run bisection search on the optimal value $t$ of problem \eqref{eq:optalpha} and check feasibility with MILP-solvers. Using the lemma to formulate MILPs corresponding to cases ($a$), ($b$), and ($c'$), we obtain the overall feasibility by sequentially checking cases ($a$), ($b$), and ($c'$) as summarized in Alg. \ref{alg:alpha}. Continue with binary search on $t$ until convergence, which corresponds to the optimal solution of $\alpha^*$ given $f_S$, $B_C$, and $S$.

\begin{algorithm}[t]
\caption{Algorithm to obtain $\alpha^*$ given $f_S$, $B_C$, and $S$}
\label{alg:alpha}
\begin{algorithmic}[1]
\renewcommand{\algorithmicrequire}{\textbf{Input:}}
\renewcommand{\algorithmicensure}{\textbf{Output:}}
\REQUIRE a small $\epsilon>0$ and a large $t^{\max}>0$.
\STATE set $t_L \leftarrow 0$ and $t_U \leftarrow t^{\max}$.
\WHILE{$t_U - t_L \geq \epsilon$}
    \STATE set $t \leftarrow (t_L + t_U)/2$, $\text{flag} \leftarrow \texttt{true}$.
    \FOR{$i \in \mathcal{I}$}
        \FOR{$j \in \mathcal{C}_i$}
            \STATE assume $j'=j$ and test feasibility of problem in Eq. \eqref{eq:optalpha} under cases (a), (b), and (c').
        \ENDFOR
        \IF{infeasible for any $j$ and (a), (b), or (c')}
            \STATE set $\text{flag} \leftarrow \texttt{false}$ and break.
        \ENDIF
    \ENDFOR
    \IF{$\text{flag}==\texttt{true}$}
        \STATE set $t_H \leftarrow t$.
    \ELSE
        \STATE set $t_L \leftarrow t$.
    \ENDIF
\ENDWHILE
\STATE obtain $\alpha^*$ corresponding to the lowest feasible $t$.
\ENSURE $\alpha^* = \{\alpha^*_i\}_{i\in\mathcal{I}}$.
\end{algorithmic}
\end{algorithm}

\subsection{Optimize $f_S$ given $\alpha$, $B_C$, and $S$}

This subproblem corresponds to finding the lowest value of $t$ and corresponding $\mathbf{f}_S$ for which problem \eqref{eq:optfs} is feasible.
\begin{align}\label{eq:optfs}
    \min_{f_S}\ 1,\ \text{s.t.: }&\ \max_iT^{\text{cluster}}_{S,i}\leq t, \notag\\
    &\ \max_iT^{\text{total}}_i\leq t,\eqref{eq:Tglob2},\eqref{eq:Tglob5}
\end{align}
It should be noted that to optimize $f_{S,i}$, by introducing $N_i$, we can reformulate Eq. \eqref{eq:Tglob5} as a cubic inequality in the variable $f_{S,i}$ featuring constant coefficients, which is efficiently reduced to disjoint intervals on the real axis. Then, we can run bisection search on the optimal value $t$ and obtain the overall feasibility by sequentially checking cases ($a$), ($b$), and ($c'$) as summarized in Alg. \ref{alg:fs}.

\begin{algorithm}[t]
\caption{Algorithm to obtain $f_S^*$ given $\alpha$, $B_C$, and $S$}
\label{alg:fs}
\begin{algorithmic}[1]
\renewcommand{\algorithmicrequire}{\textbf{Input:}}
\renewcommand{\algorithmicensure}{\textbf{Output:}}
\REQUIRE a small $\epsilon>0$ and a large $t^{\max}>0$.
\STATE set $t_L \leftarrow 0$ and $t_U \leftarrow t^{\max}$.
\WHILE{$t_U - t_L \geq \epsilon$}
    \STATE set $t \leftarrow (t_L + t_U)/2$, $\text{flag} \leftarrow \texttt{true}$.
    \FOR{$i \in \mathcal{I}$}
        \STATE determine $j'$ and the correct case from cases (a), (b), and (c') and test feasibility of problem in Eq. \eqref{eq:optfs}.
        \IF{infeasible}
            \STATE set $\text{flag} \leftarrow \texttt{false}$ and break.
        \ENDIF
    \ENDFOR
    \IF{$\text{flag}==\texttt{true}$}
        \STATE set $t_H \leftarrow t$.
    \ELSE
        \STATE set $t_L \leftarrow t$.
    \ENDIF
\ENDWHILE
\STATE obtain $f_S^*$ corresponding to the lowest feasible $t$.
\ENSURE $f_S^* = \{f_{S,i}^*\}_{i\in\mathcal{I}}$.
\end{algorithmic}
\end{algorithm}

\subsection{Optimize $B_C$ and $S$ given $\alpha$ and $f_S$}

This subproblem corresponds to finding the lowest value of $t$ and corresponding $B_C$ and $S$ for which problem \eqref{eq:optbcs} is feasible.
\begin{align}\label{eq:optbcs}
    \min_{B_C,S}\ 1,\ \text{s.t.: }&\ \max_iT^{\text{total}}_i\leq t,\notag \\
    &\ \max_iT^{\text{cluster}}_{S,i}\leq t,\eqref{eq:Tglob3},\eqref{eq:Tglob4},\eqref{eq:Tglob7}
\end{align}
Fortunately, note that Eq. \eqref{eq:Tglob4} is a jointly convex constraint in both $B_C$ and $S$, which means we can solve for the optimal $B_C$ and $S$ with one subproblem. As before, we can run bisection search on the optimal value $t$ and obtain the overall feasibility by sequentially checking cases ($a$), ($b$) and ($c'$) as summarized in Alg. \ref{alg:bcs}.

\begin{algorithm}[t]
\caption{Algorithm to obtain $B_C^*$ and $S^*$ given $\alpha$ and $f_S$}
\label{alg:bcs}
\begin{algorithmic}[1]
\renewcommand{\algorithmicrequire}{\textbf{Input:}}
\renewcommand{\algorithmicensure}{\textbf{Output:}}
\REQUIRE a small $\epsilon>0$ and a large $t^{\max}>0$.
\STATE set $t_L \leftarrow 0$ and $t_U \leftarrow t^{\max}$.
\WHILE{$t_U - t_L \geq \epsilon$}
    \STATE set $t \leftarrow (t_L + t_U)/2$, $\text{flag} \leftarrow \texttt{true}$.
    \FOR{$i \in \mathcal{I}$}
        \STATE determine $j'$ and test feasibility of problem in Eq. \eqref{eq:optbcs} under cases (a), (b), and (c').
        \IF{infeasible for any of cases (a), (b), or (c')}
            \STATE set $\text{flag} \leftarrow \texttt{false}$ and break.
        \ENDIF
    \ENDFOR
    \IF{$\text{flag}==\texttt{true}$}
        \STATE set $t_H \leftarrow t$.
    \ELSE
        \STATE set $t_L \leftarrow t$.
    \ENDIF
\ENDWHILE
\STATE obtain $B_C^*$ and $S^*$ corresponding to the lowest feasible $t$.
\ENSURE $B_C^* = \{B_{C,i}^*\}_{i\in\mathcal{I}}$ and $S^*=\{S_i\}_{i\in\mathcal{I}}$.
\end{algorithmic}
\end{algorithm}

\section{Hierarchical FL Problem Formulation}\label{sec:hfl}

The proposed $\mathrm{ParaBlock}$++ system leverages NTNs, which consist of LEO constellations for global coverage to aggregate, broadcast, and propagate models. The entire process comprises $T$ global (or cloud-level) rounds of model aggregation at the cloud server. Each global round comprises $E$ cluster (or edge-level) rounds of client submodel aggregation at the edge server. Each cluster round comprises $H$ local (or client-level) training rounds, during which a client performs submodel updates using SGD (or other optimizers) on a mini-batch of client-local data.

In addition, we consider a proper partitioning of the model to leverage BCD-type update, which significantly accelerates the convergence of training. To mitigate the inherent issues of sequential updates and the lack of parallelism with BCD-type updates, our integrated model interleaves training and communication to improve concurrency.

\subsection{Objective Functions}

Given model parameter $x$ and data sample $\xi$, we decompose the objective function by $\phi(x;\xi)=f(x;\xi)+r(x)$ \cite{bsg}, where $f(x;\xi)$ denotes the empirical loss and $R(x)=\sum_{i\in\mathcal{I}}r_i(x_i)$ denotes the structural loss (possibly non-differentiable regularization independent from $\xi$), where $x_i$ is the portion of $x$ associated with block $i$. Given $f(x)$ and $R(x)$, we construct empirical loss functions at the client, cluster, and global levels.
\begin{align*}
    f_j(x)=&\ \mathbb{E}_{\xi\sim\bar{\mathcal{D}}_j}[f(x;\xi)],\ \forall x,\ \forall j\in\mathcal{C}_i,\ \tag{client loss}\\
    f_i(x)=&\ \mathbb{E}_{\xi\sim\mathcal{D}_i}[f(x;\xi)],\ \forall x,\ \forall i\in\mathcal{I},\ \tag{cluster loss}\\
    F(x)=&\ \mathbb{E}_{\xi\sim\mathcal{D}}[f(x;\xi)],\ \forall x,\ \tag{global loss}
\end{align*}
where $\mathcal{D}_j$ and $\bar{\mathcal{D}}_j\!\triangleq\!\mathcal{D}_j\backslash\bar{\mathcal{D}}_j^s$ denote the original dataset of client $j$ and the dataset after offloading insensitive data to server $i$, resp. $\mathcal{D}_i\!\triangleq\!\sum_{j\in\mathcal{C}_i}\mathcal{D}_j$ denotes the union of datasets in cluster $i$, and $\bar{\mathcal{D}}_i\triangleq\sum_{j\in\mathcal{C}_i}\bar{\mathcal{D}}_j^s$ denotes non-sensitive data offloaded from clients $j\in\mathcal{C}_i$ to server $i$, which originally has no local dataset. $\mathcal{D}$ denotes the virtual dataset that represents ground-truth data distribution.

We denote the update of $x_i$ with client $j$ at iteration $(t,e,h)$ by $x^{t,e,h}_{i,j}$, where $t\in[T], e\in[E], h\in[H], i\in\mathcal{I}, j\in\mathcal{C}_i$. The cloud server initiates the training process by solving the optimization problems in Section \ref{sec:iterative} and partitions model $S$ into submodels $\{S_i\}_{i\in\mathcal{I}}$, whose parameters $x_i$ are distributed to the corresponding edge servers. Without loss of generality, assume that edge server $i$ receive $x^{t}_i$ and have it broadcast to clients $j\in\mathcal{C}_i$. That is, all clients in cluster $i$ obtain identical submodel $S_i$ with parameters $x^{t,0,0}_{i,j}=x^{t}_i$ and begin to train $S_i$ with their local dataset $\bar{\mathcal{D}}_j$. In addition, we assume that the server acts as an ordinary client when training the submodel, using dataset $\bar{\mathcal{D}}_i$.

\subsection{Hierarchical Update Rules}

\noindent\textit{1) Client update}. The objective function for client $j\in\mathcal{C}_i$ is defined with a mini-batch $\Theta^{t,e,h}_{i,j}$ of local data at client $j$ as
\begin{align}\label{eq:update1}
    \phi_{i,j}(x^{t,e,h}_{i,j})=\mathbb{E}_{\xi\in\Theta^{t,e,h}_{i,j}}[\phi(x^{t,e,h}_{i,j};\xi)],\ \forall t,e,h,\ \forall i,j
\end{align}
where client $j$ computes stochastic gradients with respect to submodel parameters ${x}^{t,e,h}_{i,j}$ and updates it with learning rate $\alpha_t$ for global iteration $t$,
\begin{align}\label{eq:update2}
    {x}^{t,e,h+1}_{i,j}={x}^{t,e,h}_{i,j}-\alpha^t\cdot\nabla_{x_i}\phi_{i,j}({x}^{t,e,h}_{i,j}),\ \forall t,e,h,\ \forall i,j.
\end{align}

\noindent\textit{2) Cluster update}. At the end of the $H$-th local update, each client $j$ uploads submodel ${x}^{t,e,H}_{i,j}$ to edge server $i$. The edge server $i$ aggregates the client submodels with
\begin{align}\label{eq:update3}
    {x}^{t,e+1}_{i}\!=\!\frac{1}{|{\mathcal{D}}_i|}\Big(\textstyle\sum_{j\in\mathcal{C}_i}|\bar{\mathcal{D}}_j|{x}^{t,e,H}_{i,j}\!+\!|\bar{\mathcal{D}}_i|{x}^{t,e,H}_{i,i}\Big), \forall t,e, \forall i
\end{align}
and broadcasts the aggregated submodel back to clients for subsequent cluster round $e+1$.

\noindent\textit{3) Global update}. After the $E$-th cluster update, edge server $i$ transmits the updated submodel to the edge server in charge of cluster $(i\!\!\mod s)+1$. This stands in sharp contrast with conventional HFL frameworks, where edge server $i$ uploads the updated submodel to the cloud server.


\section{Convergence Analysis}\label{sec:convergence}

In this section, we analyze the convergence of BCD-type updates for composite objective functions $\Phi(x)=F(x)+R(x)$ of varying convexity. Without loss of generality, we assume the update order $\pi(i)=i$ for all $i\in\mathcal{I}$ in Alg. \ref{alg:parablock++}.

\subsection{Notations}

For conciseness, let $g^{t,e,h}_{i,j}$ denote the full-batch gradient at client $j$ with dataset $\bar{\mathcal{D}}_j$ in cluster $i$ at iteration $(t,e,h)$, $\tilde{g}^{t,e,h}_{i,j}$ denote the corresponding mini-batch stochastic gradient, and $\delta^{t,e,h}_{i,j}=\tilde{g}^{t,e,h}_{i,j}-{g}^{t,e,h}_{i,j}$ denote the difference of them. Unlike traditional FL, the BCD-type updates are explicitly given by
\begin{align*}
    g^{t,e,h}_{i,j}=\nabla_{x_i}f_j([x^{t+1,0,0}_{<i},x^{t,e,h}_{\geq i}]),\ \forall t,e,h,\ \forall i,j\\
    g^{t,e,h}_i=\nabla_{x_i}f_i([x^{t+1,0,0}_{<i},x^{t,e,h}_{\geq i}]),\ \forall t,e,h,\ \forall i,j\\
    G^{t,e,h}_i=\nabla_{x_i}F([x^{t+1,0,0}_{<i},x^{t,e,h}_{\geq i}]),\ \forall t,e,h,\ \forall i,j,
\end{align*}
implying that in any global round $t$, block $i$ is updated after all blocks $<i$ have been updated, which significantly complicates the proof of convergence.

\subsection{Assumptions}

Our theoretical analysis for convergence focuses on convex, strongly convex, and smooth and non-convex functions, resp., under general non-i.i.d. dataset distributions. Here, we list all assumptions that are consistent with the FL literature.\\

\noindent\textbf{Assumption 1} (Lower Boundedness). The global loss function is lower bounded, i.e., $\Phi(x) > -\infty,\ \forall x$.

\noindent\textbf{Assumption 2} (L-Smoothness and L-Continuity). The global empirical loss function is differentiable and $L$-smooth, i.e.,
\begin{align*}
    \|\nabla_{x_i}F(x)-\nabla_{x_i}F(y)\|\leq L\|x-y\|,\ \forall x,y,\ \forall i.
\end{align*}
The structural loss functions $r_i$ are $L_{r_i}$-continuous, i.e.,
\begin{align*}
    \|r_i(x_i)-r_i(y_i)\|\leq L_{r_i}\|x_i-y_i\|,\ \forall x_i,y_i,\ \forall i.
\end{align*}
We let $L_r \triangleq \max_i L_{r_i}$ be the dominant Lipschitz constant.

\noindent\textbf{Assumption 3} (Unbiased Stochastic Gradient). For dataset $\bar{\mathcal{D}}_j$, the stochastic gradient with mini-batch $\Theta^{t,e,h}_{i,j}\!\subset\!\bar{\mathcal{D}}_j$ provides an unbiased estimator of the true (i.e., full-batch) gradient, i.e.,
\begin{align*}
    \mathbb{E}_{\Theta^{t,e,h}_{i,j}\subset\bar{\mathcal{D}}_j}[\tilde{g}^{t,e,h}_{i,j}(\Theta^{t,e,h}_{i,j})]=g^{t,e,h}_{i,j},\ \forall t, e, h,\ \forall i, j.
\end{align*}

\noindent\textbf{Assumption 4} (Bounded Stochastic Gradient Variance). For dataset $\bar{\mathcal{D}}_j$, the variance of the stochastic gradient with mini-batch $\Theta^{t,e,h}_{i,j}\subset\bar{\mathcal{D}}_j$ is bounded, i.e.,
\begin{align*}
    \mathbb{E}_{\Theta^{t,e,h}_{i,j}\subset\bar{\mathcal{D}}_j}\big[\big\|\tilde{g}^{t,e,h}_{i,j}(\Theta^{t,e,h}_{i,j})-g^{t,e,h}_{i,j}\big\|^2\big]\leq(\sigma^t)^2,\ \forall t, e, h,\ \forall i, j.
\end{align*}

\noindent\textbf{Assumption 5} (Bounded Intra-Cluster Gradient Variance). The gradient dissimilarity of the empirical loss functions between client $j$ and cluster $i$ is bounded, i.e.,
\begin{align*}
    \|g_{i,j}^{t,e,h}-g_i^{t,e,h}\|^2\leq(\delta^t_1)^2,\ \forall t,e,h,\ \forall i,j.
\end{align*}

\noindent\textbf{Assumption 6} (Bounded Inter-Cluster Gradient Variance). The gradient dissimilarity of the empirical loss functions between cluster $i$ and the truth is bounded, i.e.,
\begin{align*}
    \|g_{i}^{t,e,h}-G^{t,e,h}_i\|^2\leq(\delta^t_2)^2,\ \forall t,e,h,\ \forall i.
\end{align*}

\noindent\textbf{Assumption 7} There exists a constant $A>0$ such that for the learning rate $\alpha^t$ we choose,
\begin{align*}
    \|\mathbb{E}[\delta^{t,e,h}_{i,j}]\|\leq A\alpha^t,\ \forall t,e,h,\ \forall i,j.
\end{align*}

\noindent\textbf{Assumption 8} There exists a constant $\rho$ such that
\begin{align*}
    \mathbb{E}[\|x^{t,e,h}_{i,j}\|^2]\leq\rho^2,\ \forall t,e,h,\ \forall i,j.
\end{align*}

Assumptions 1-4 have been widely adopted in the context of stochastic non-convex optimization. Assumptions 5 and 6 characterize the degree of data heterogeneity across clusters and clients within the HFL literature. In addition, Assumptions 7 and 8 are required for BCD-type updates \cite{bsg}.




\begin{algorithm}[H]
\caption{Training Procedure for $\mathrm{ParaBlock}$++}
\label{alg:parablock++}
\begin{algorithmic}[1]
\renewcommand{\algorithmicrequire}{\textbf{Input:}}
\renewcommand{\algorithmicensure}{\textbf{Output:}}
\REQUIRE servers $\mathcal{I}$, clients $\mathcal{J}$, local rounds per cluster round $H$, cluster rounds per global round $E$, global rounds $T$, initial model $x$, submodels $\{S_i\}_{i\in\mathcal{I}}$, learning rate $\alpha^t$.
\FOR{$t = 0, 1, \ldots, T-1$}
  \STATE Generate shuffled order $\pi(i),\ \forall i\in\mathcal{I}$.
  \FOR{$e = 0, 1, \ldots, E-1$}
    \FOR {$h = 0, 1, \ldots, H-1$}
      \STATE Clients $j\in\mathcal{C}_i$ update submodel parameter $x^{t,e,h}_{i,j}$ using Eq. \eqref{eq:update2} in parallel. \\
    \ENDFOR
    \STATE Edge servers update submodel parameter $x_i^{t,e+1}$ using Eq. \eqref{eq:update3} in parallel. \\
    \STATE Edge servers broadcast the updated submodel parameter $x_i^{t,e+1}$ to clients in parallel. \\
  \ENDFOR
  \FOR {$i\in\mathcal{I}$}
    \STATE Edge server $i$ transmits submodel parameter $x_i^{t}$ to edge server $(i\!\!\mod s)+1$ in parallel. \\
    \STATE Edge server $(i\!\!\mod s)+1$ broadcasts submodel parameter $x_i^t$ to clients $j\in\mathcal{C}_{(i\!\!\mod s)+1}$. \\
  \ENDFOR
\ENDFOR
\ENSURE Optimized $x^{T}$.
\end{algorithmic}
\end{algorithm}

\subsection{Convergence Results}

\noindent\textbf{Theorem 1} (Convergence for general convex functions). Let $\{x^t\}$ be generated from $\mathrm{ParaBlock}$++ with $\alpha^t={\theta}/{\sqrt{t+1}}<{1}/{L}$, $\forall t$, for some constant $0<\theta<{1}/{L}$. Under Assumptions 1 through 8, if $F$ and $r_i$'s are all convex, $x^*$ is the optimal solution and $\sigma=\sup_t\sigma^t<\infty$, then we have
\begin{align*}
    \mathbb{E}\big[\Phi(\tilde{x}^T)-\Phi(x^*)\big]\leq \frac{\theta(\ln T+1)}{\sqrt{T+1}}D(x^*)+\frac{\|x^*-x^0\|^2}{2EH\theta\sqrt{T+1}},
\end{align*}
which coincides with the well-known optimal convergence rate $\mathit{O}(1/\sqrt{t})$ under general convex settings for SGD.\\

\noindent\textbf{Theorem 2} (Convergence for strongly convex functions). Let $\{x^t\}$ be generated from $\mathrm{ParaBlock}$++ with $\alpha_t\! =\! {\theta}/{(t+1)}\! <\! {1}/{L}, \ \forall t$. Under Assumptions 1 through 8, if $F$ and $r_i$'s are convex, $\Phi$ is strongly convex with modulus $\mu > 0$, $x^*$ is the optimal solution, and $\sigma = \sup_t \sigma^t < \infty$, then for $T\geq(\lfloor A\rfloor+1)/(EH)$ with $A={EH\mu\theta}/{(2+2\mu\theta)}$, we have
\begin{align*}
    &\mathbb{E}[\|x^T-x^*\|^2]\leq\frac{c}{T},
\end{align*}
where
\begin{align*}
    &c= \begin{cases}4EHD(x^*)\theta^2/(A-1)\ (A>1), \\ \max \big\{2EHD(x^*)\theta^2 / A, \mathbb{E}[\|x^1-x^*\|^2]\big/(EH)\} \ (A \leq 1),\end{cases}
\end{align*}
which coincides with the well-known optimal convergence rate $\mathit{O}(1/t)$ under strongly convex settings for SGD.\\

\noindent\textbf{Theorem 3} (Convergence for non-convex functions). Let $\{x^t\}$ be generated from $\mathrm{ParaBlock}$++ with $r_i=0$ and $\mathcal{X}_i=\mathbb{R}^{n_i}$ for all $i$ and with positive scalars $\alpha^t$ that satisfy $\alpha^t \geq \alpha^{t+1}$ for all $t$, and $\textstyle\sum_{t=0}^{\infty}\alpha^t=+\infty,\ \sum_{t=0}^{\infty}(\alpha^t)^2<+\infty$. Under Assumptions 1 through 8, if $\sigma=\sup_t \sigma^t<\infty$ then we have
\begin{align*}
    \lim_{t\to\infty}\mathbb{E}[\|g^{t,e,h}_{i,j}\|]=0,\ \forall e, h,\ \forall i, j,
\end{align*}
which ensures vanishing gradient norm $\mathit{o}(1)$.

We summarize the main steps of the $\mathrm{ParaBlock}$++ algorithm in Alg. \ref{alg:parablock++}. Note that the propagation stage, which occurs concurrently with training, is not included for clarity.

\begin{table*}[!t]
\centering
\caption{Simulation Settings for the LEO Megaconstellation}
\label{tab:simulationsetting}
\begin{tabular}{|c|c||c|c||c|c|}
\hline
\textbf{Parameter} & \textbf{Value} & \textbf{Parameter} & \textbf{Value} & \textbf{Parameter} & \textbf{Value} \\
\hline
$f_{S,i}$ & $1$ GHz & $m_{S,i}\ ^\dagger$ & $3 \times 10^7$ cycles/sample & $P_{S,i}$ & $10$ W \\
$f_{C,j}$ & $3 \times 10^8$ Hz & $m_{C,j}\ ^\dagger$ & $3 \times 10^7$ cycles/sample & $P_{C,j}$ & $0.3$ W \\
$T$ & $360$ sec & $\xi$ & $2$ & $N_0$ & $3.98 \times 10^{-21}$ W/Hz \\
$\kappa$ & $10^{-28}$ Joules$\cdot$seconds$^2$/cycles$^3$ & $B_i$ & $10$ MHz & $-$ & $-$ \\
\hline
\end{tabular}\\
$\ ^\dagger$: This estimate applies to the full ResNet-18 model and scales proportionally with the submodel size, i.e., as a fraction of the total parameters.
\end{table*}

\section{Experimental Results}\label{sec:results}

Our proposed $\mathrm{ParaBlock}$++ framework consists of a LEO satellite communication model and an HFL model. We conduct separate experiments for both components, and combine the results to demonstrate model performance versus training and communication time. For the LEO satellite communication model, we adopt satellite hyperparameters from previous research \cite{gts} and Starlink's specification in its official filing \cite{starlink}. For the HFL model, we use models having objectives of varying convexity. All experiments were conducted on a dedicated workstation with an Intel i9-14900KF/RTX 4090 GPU workstation with 32 GB RAM, using Ubuntu 20.04-LTS and Python 3.11.14. All model training scripts are implemented using PyTorch 2.6.0 \cite{pytorch}.

\subsection{Satellite Megaconstellation Simulation Setup}

We simulate LEO satellite megaconstellations with technical specifications from the recent Starlink official release of FCC approval \cite{starlink}. We employ the 550km shell, which has 72 orbit planes times 22 equally spaced satellites per each at an altitude of 550km and an inclination of $53^\circ$. Using MATLAB built-in \texttt{walkerStar} \cite{walkerstar} and \texttt{orbitalElements} \cite{orbitalelements} functions, we obtain the Starlink megaconstellations and orbital parameters for individual LEO satellites that are required to construct the satellite routing graphs.

Since transmission parameters for Starlink are not officially released, we use the simulation settings proposed in a recent research \cite{gts} as summarized in Table. \ref{tab:simulationsetting}. However, we discard the transmission rate for satellite handover $R^{\text{ISL}}=3.125$ Mbps since it is significantly lower than that achieved with Starlink \cite{starlink}. Instead, we recompute the ISL and GSL rates using the satellite megaconstellation parameters of Starlink \cite{starlink}.

\subsection{Hierarchical Federated Learning Implementation}

In this subsection, we briefly introduce the datasets, classification models, and baselines.

\textit{1) Datasets}. The FashionMNIST is a multi-class image classification dataset for benchmarking convolutional neural networks (CNNs) and other deep learning models such as ResNet-18. The dataset contains 60000 training images and 10000 test images, with each 28x28 grayscale images representing one of 10 distinct categories of clothing items, maintaining the same format as MNIST. We conduct experiments under both data homogeneity (i.e., i.i.d.) and heterogeneity (i.e., non-i.i.d.), as controlled by the inter-cluster and intra-cluster concentration parameters of the Dirichlet distribution. We assign samples to cluster $i\in\mathcal{I}$ according to a Dirichlet distribution $\mathrm{Dir}(\alpha)$, where the concentration parameter $\alpha>0$ governs the degree of data heterogeneity, with smaller values of $\alpha$ inducing more non-i.i.d. distributions.

\textit{2) Classification Models}. We select classifiers with varying convexity, including the non-regularized convex logistic regression (LR) \cite{lr} (convex but not strongly convex), the regularized Crammer-Singer SVM \cite{crammersinger}, and the 18-layer residual network (ResNet-18) \cite{resnet18} (non-convex). The LR model has 28x28x10 weights and 10 biases with regularization coefficient $C=0.0$, the Crammer-Singer SVM has 28x28x10 weights and 10 biases with regularization coefficient $C=1.0$, and the ResNet-18 model is a lightweight choice (having about 11.7M parameters) in the ResNet model series, which has a non-convex objective function as a direct consequence of modular design with DNNs.

\textit{3) Baselines}. In federated learning across geographically distributed clusters, hierarchical aggregation is essential for communication efficiency. Building on this HFL structure, we incorporate two advanced baselines: (i) hierarchical independent submodel training (HIST) \cite{hist}, where each cluster trains its submodel concurrently and the cloud aggregates uploaded edge updates; and (ii) traditional federated BCD (FedBCD), whose basic variant \cite{fedbcd5} sequentially trains one submodel at a time across all clusters, with the cloud averaging the resulting updates.

In the remainder of this section, we experiment with various combinations of hyperparameters, including global rounds $T$, cluster rounds $E$, client rounds $H$, number of blocks $s=|\mathcal{I}|$, clients per cluster $\{|\mathcal{C}_i|\}_{i\in\mathcal{I}}$, inter-cluster heterogeneity $\alpha_\text{inter}$, intra-cluster heterogeneity $\alpha_\text{intra}$, and learning rate $\alpha^t$, to have comprehensive performance metrics. We emphasize that the batch size $b$ is taken to be 32 throughout all evaluations.

\subsection{Accuracy versus Rounds}

\begin{figure}[htbp]
    \vspace{-10pt}
    \includegraphics[width=0.24\textwidth]{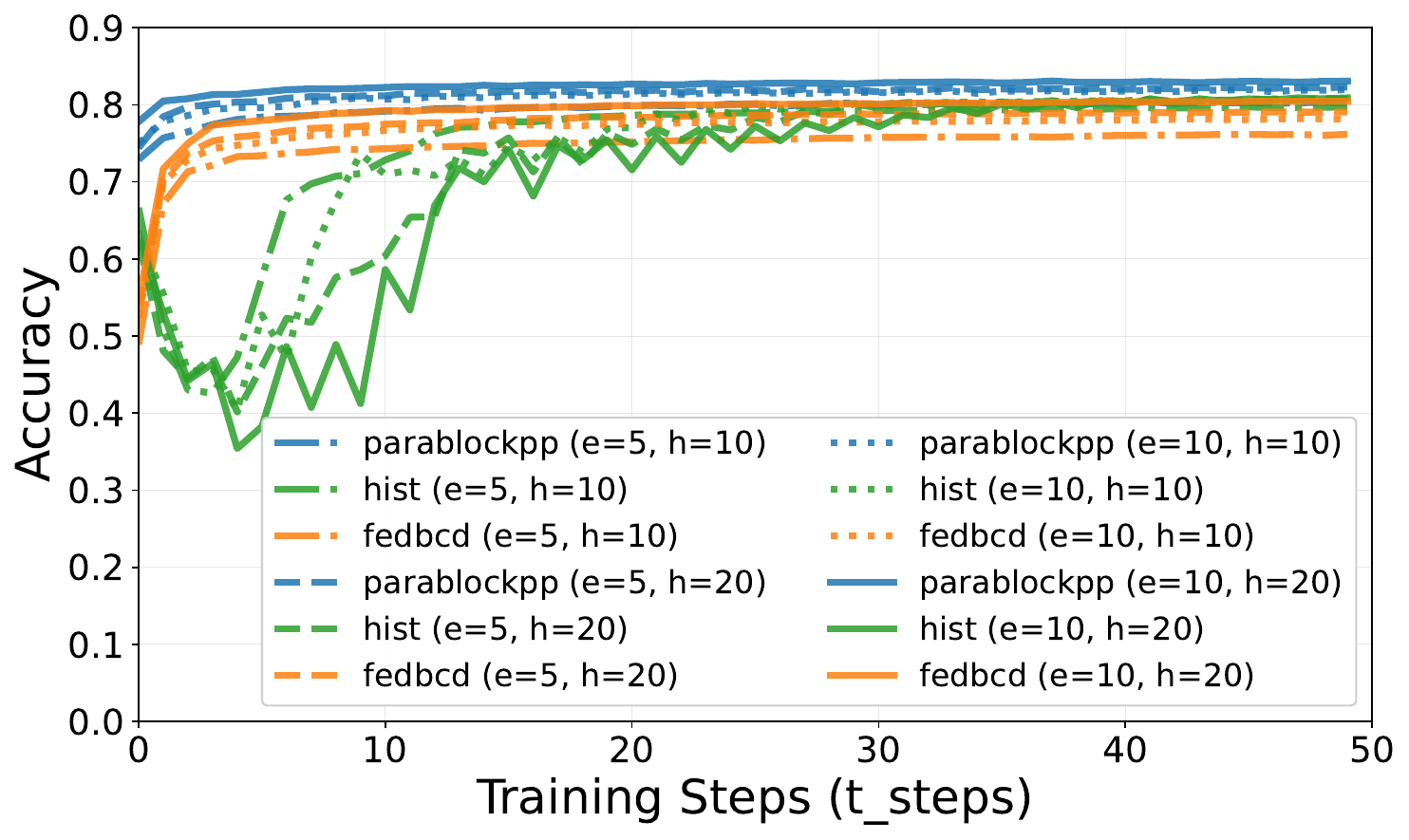}
    \includegraphics[width=0.24\textwidth]{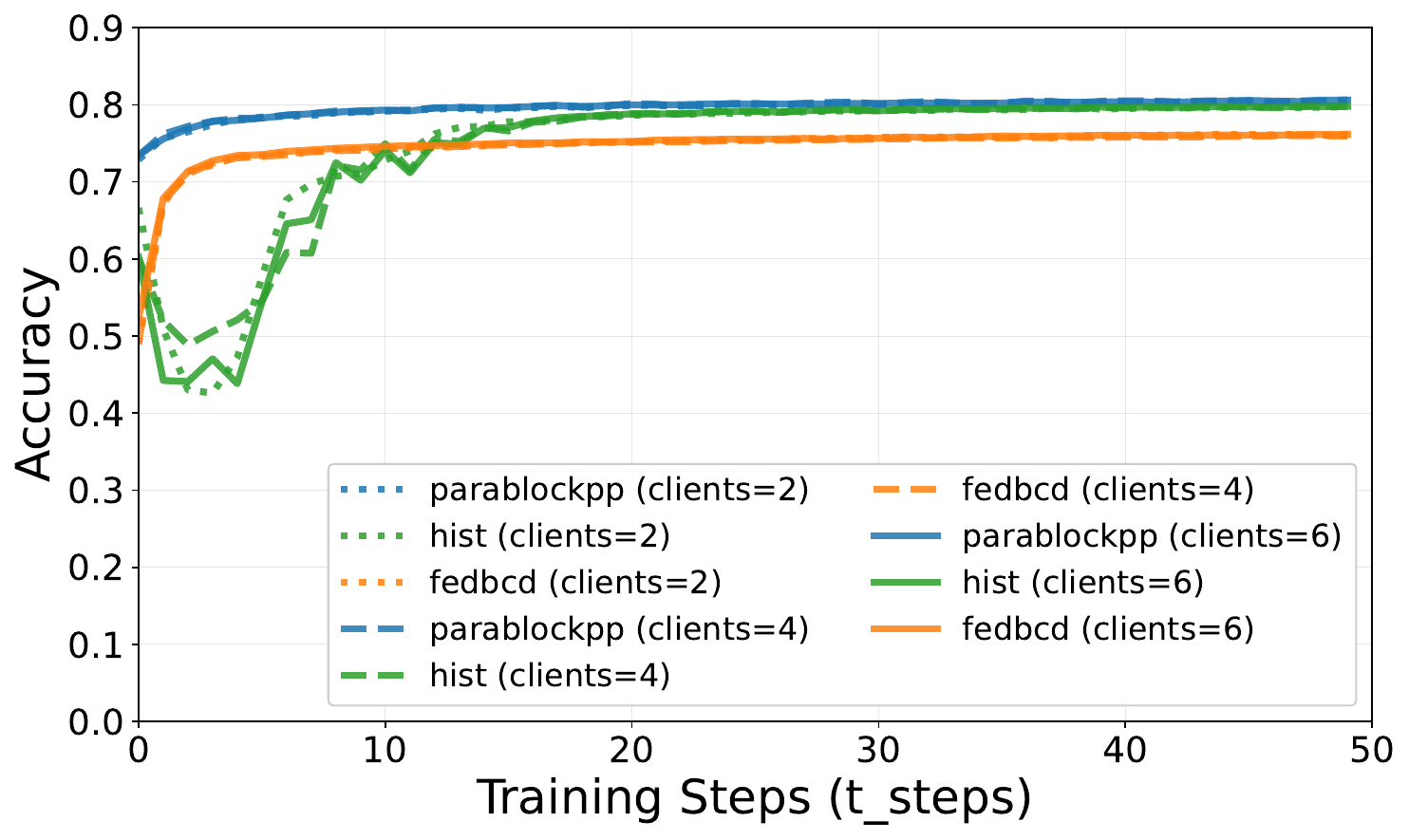}
    \vspace{10pt}
    \includegraphics[width=0.24\textwidth]{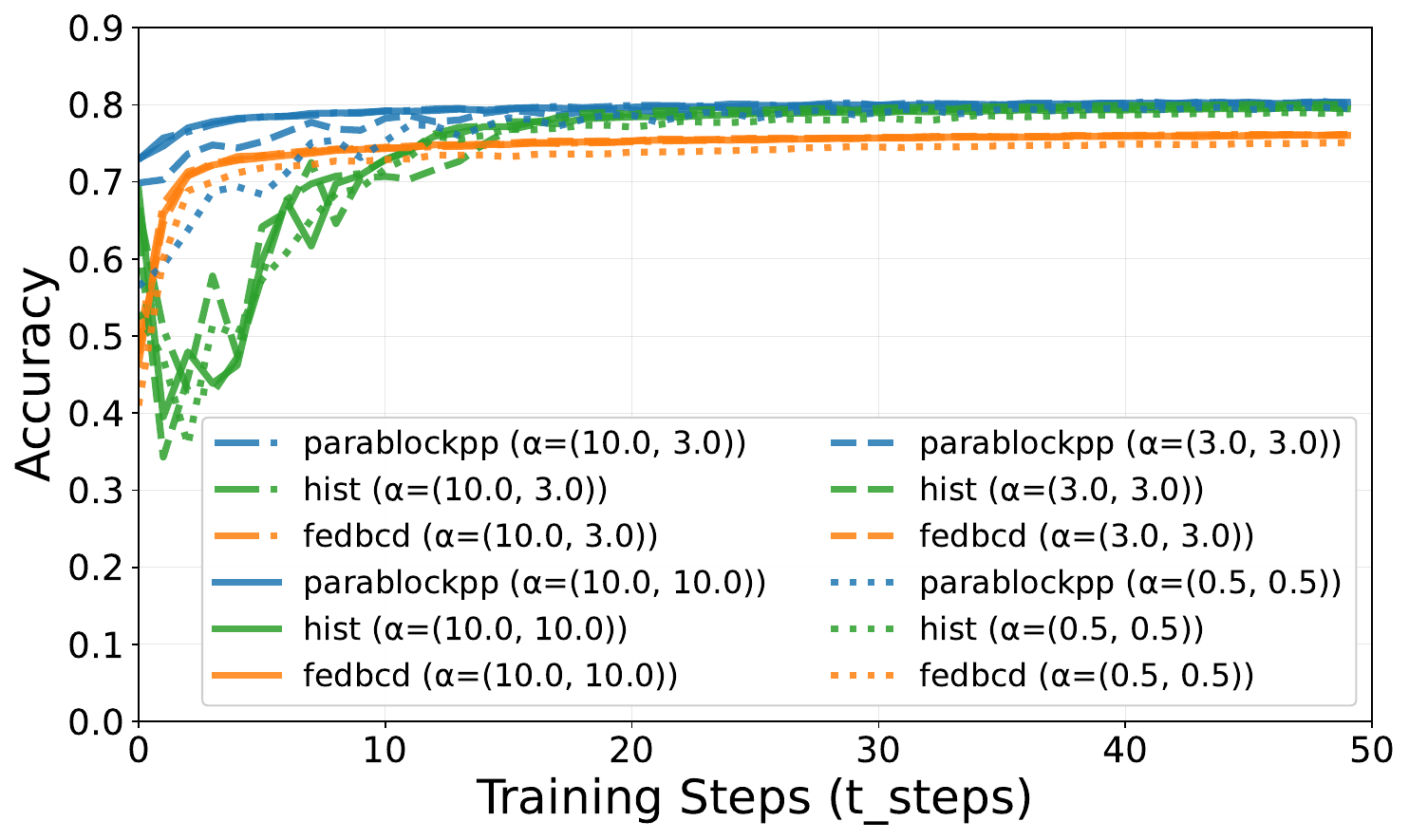}
    \includegraphics[width=0.24\textwidth]{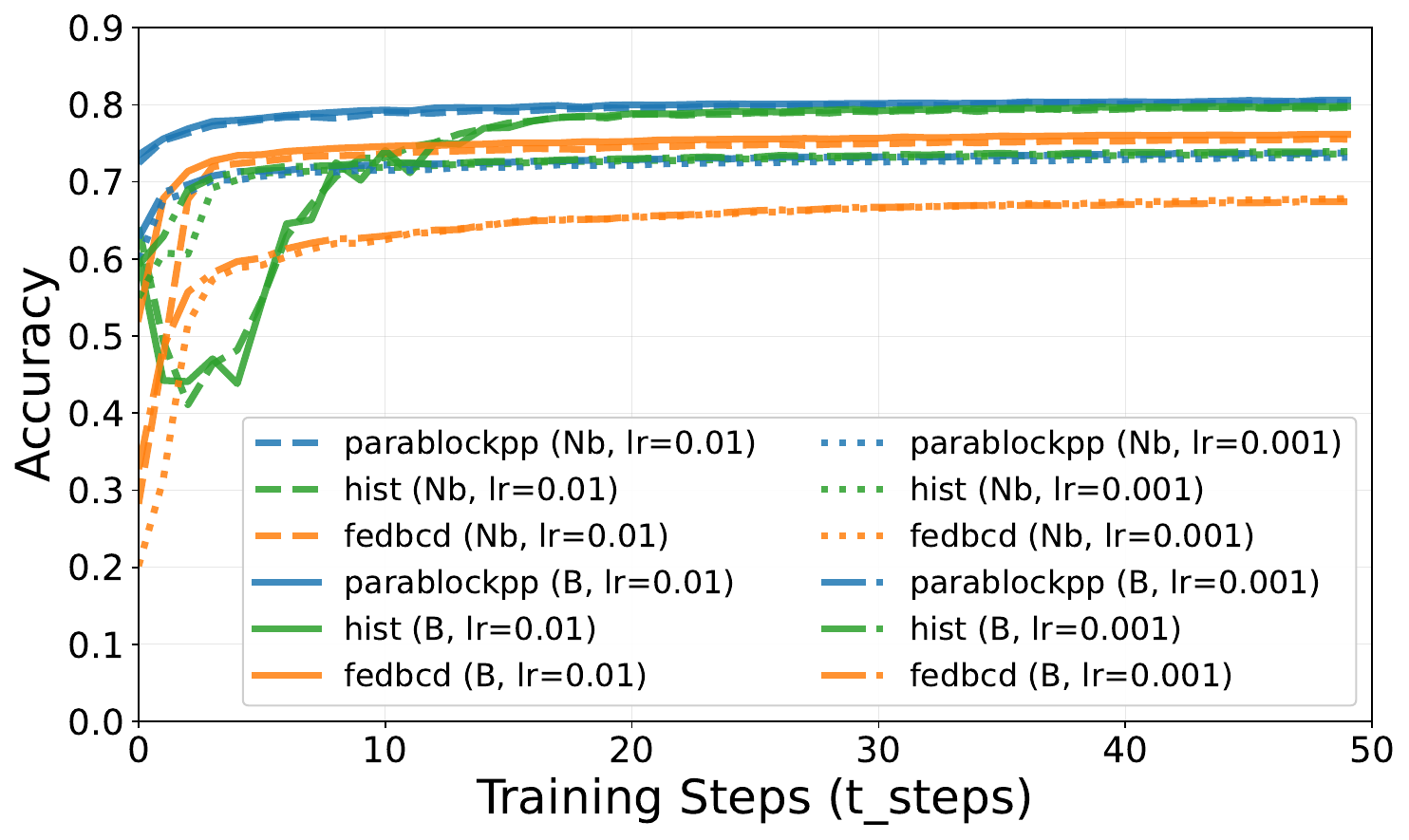}
    \vspace{-30pt}
    \caption{Comparison for LR with default hyperparameters: balanced partition, $T\!=\!50$, $E\!=\!5$, $H\!=\!10$, $\#\text{clusters}\!=\!4$, $\#\text{clients}\_\text{per}\_\text{cluster}\!=\!2$, $\alpha_\text{inter}\!=\!10.0$, $\alpha_\text{intra}\!=\!3.0$, $\alpha^t\!=\!0.01/(t+1)$ and batch size $b\!=\!32$. Upper left: varying $E$ and $H$, upper right: varying $\#\text{clients}\_\text{per}\_\text{cluster}$, lower left: varying $\alpha_\text{inter}$ and $\alpha_\text{intra}$, lower right: varying partition balance and $\alpha^t$.}
    \label{fig:lr}
    \vspace{-10pt}
\end{figure}

\begin{figure}[htbp]
    \vspace{-10pt}
    \includegraphics[width=0.24\textwidth]{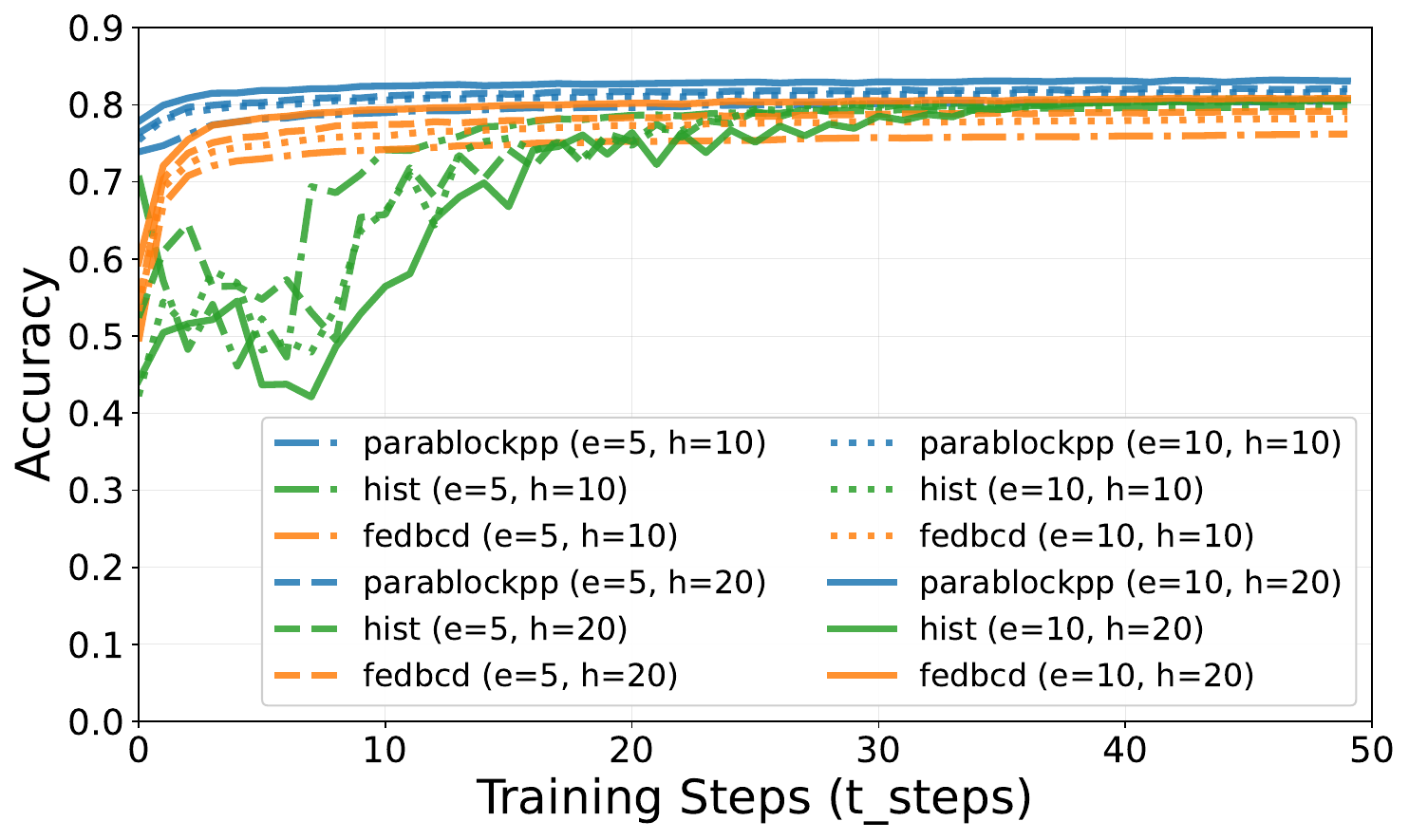}
    \includegraphics[width=0.24\textwidth]{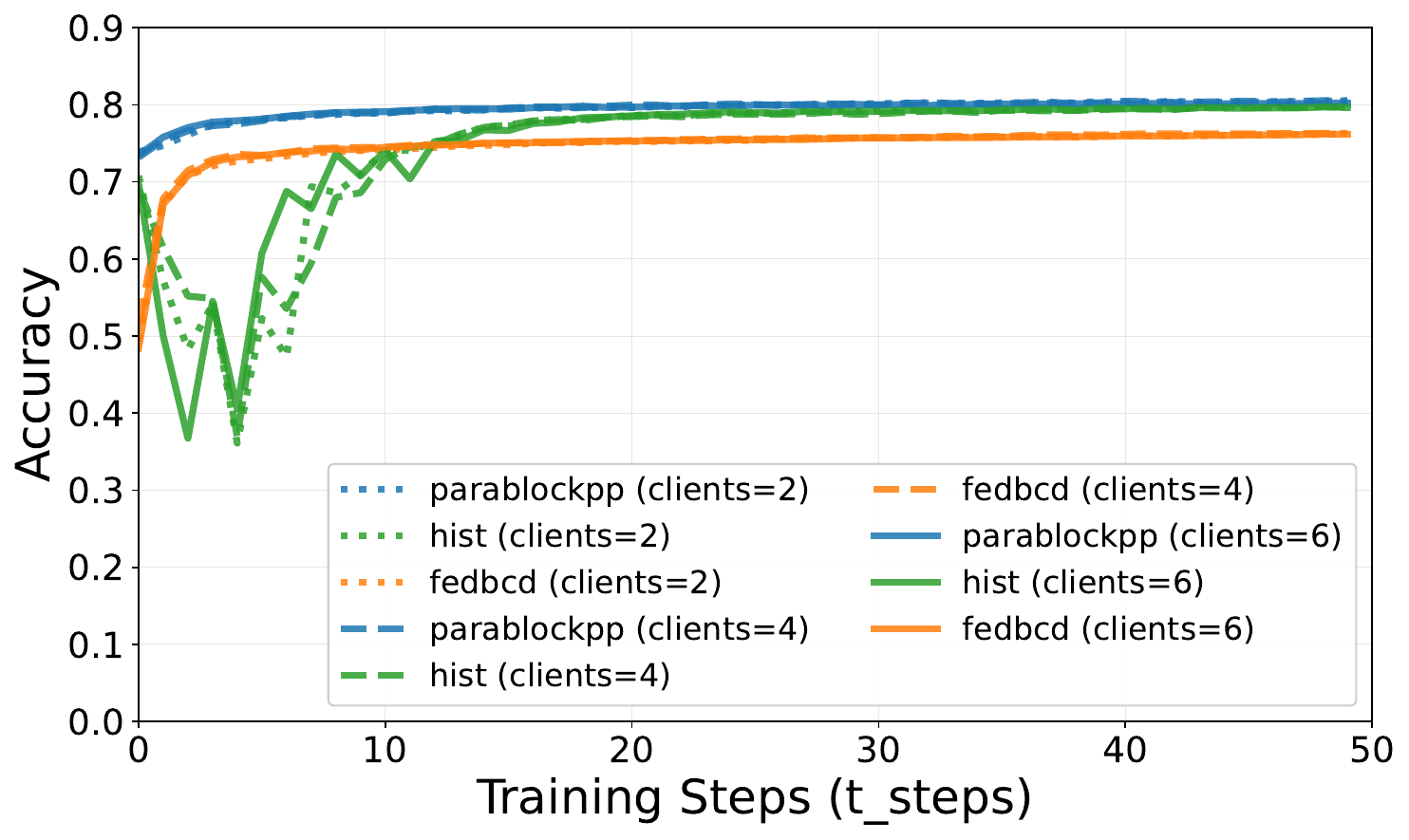}
    \vspace{10pt}
    \includegraphics[width=0.24\textwidth]{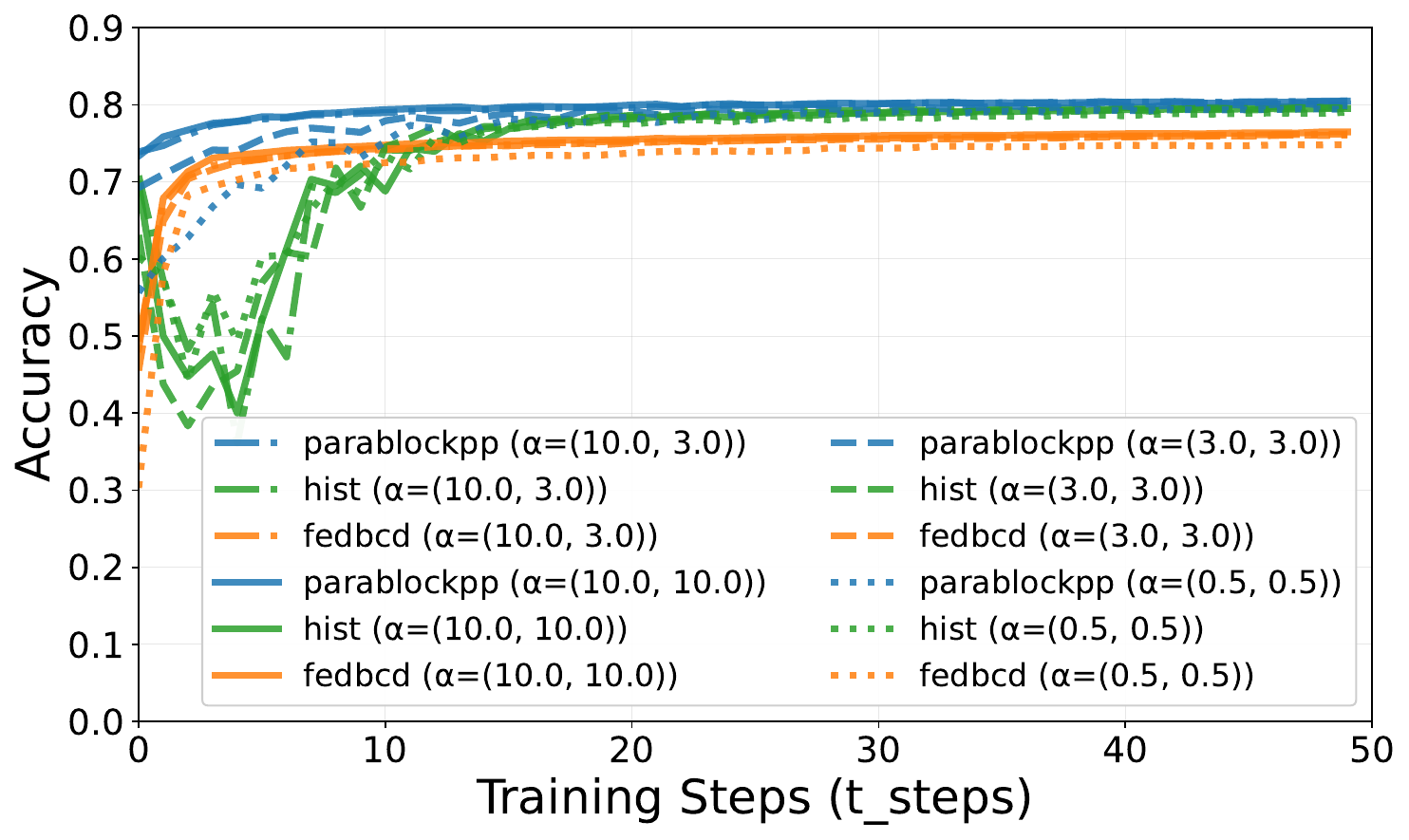}
    \includegraphics[width=0.24\textwidth]{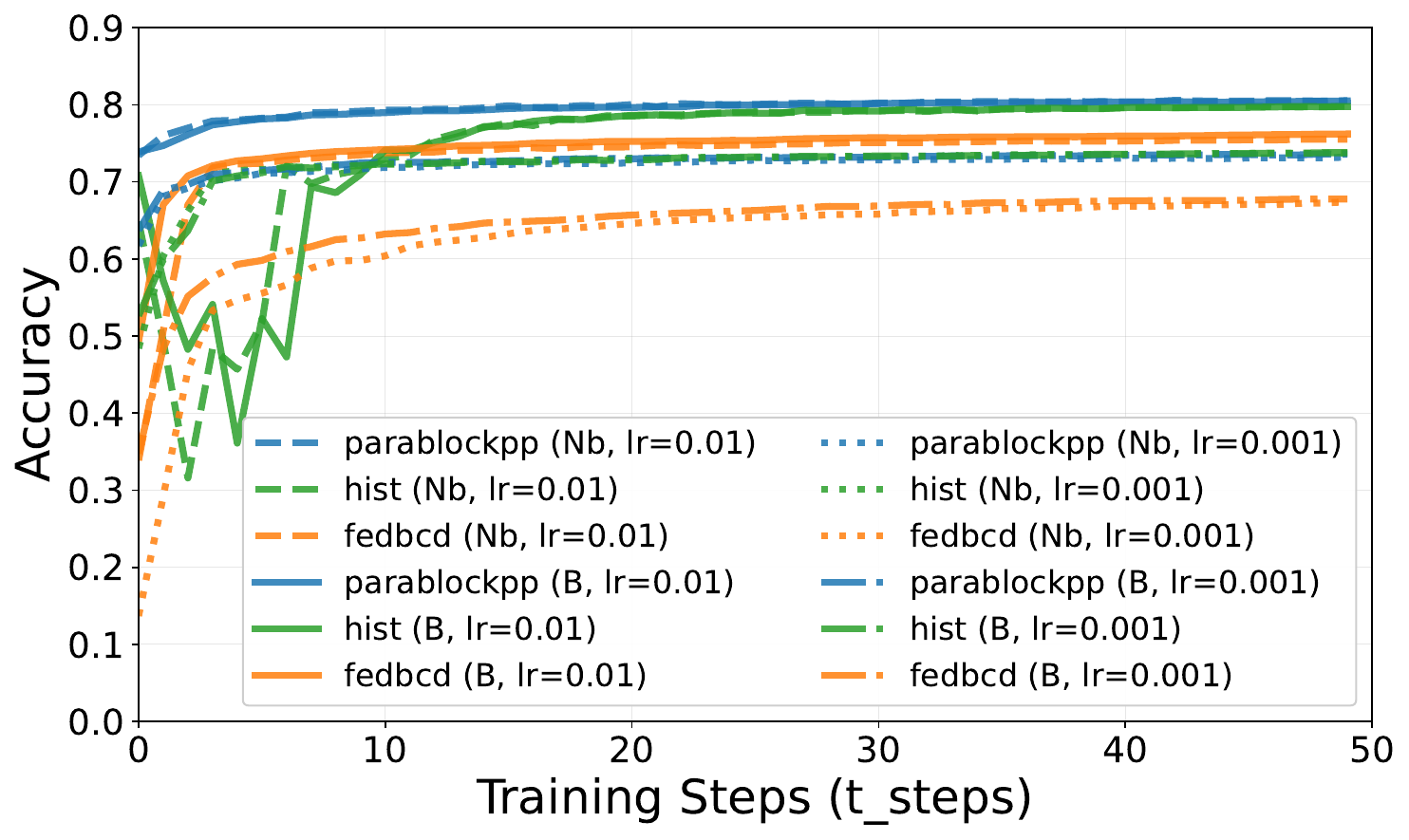}
    \vspace{-30pt}
    \caption{Comparison for Crammer-Singer SVM with the same default parameters and subplots as for LR.}
    \label{fig:crammersingersvm}
    \vspace{-10pt}
\end{figure}

\begin{figure}[htbp]
    \vspace{-10pt}
    \includegraphics[width=0.24\textwidth]{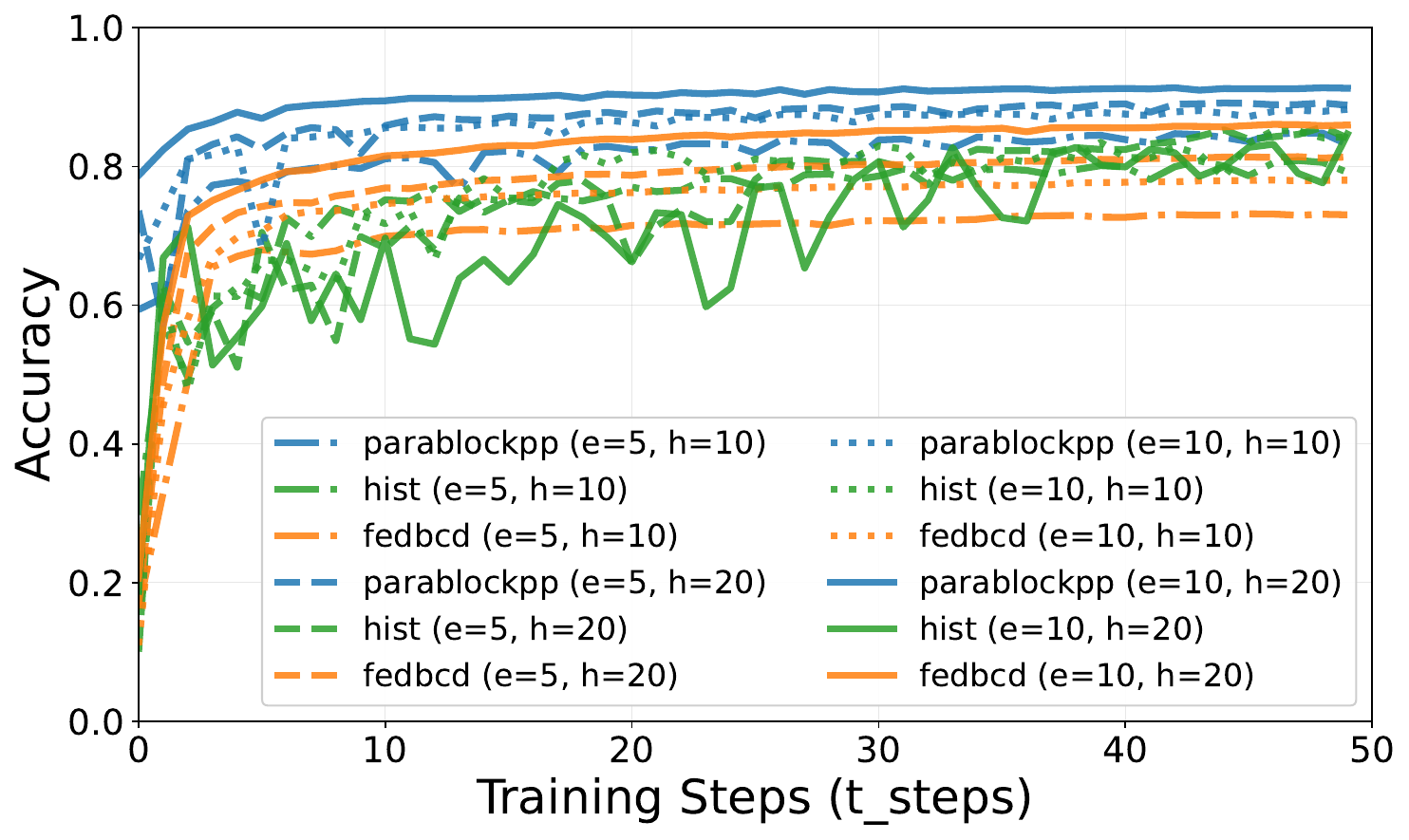}
    \includegraphics[width=0.24\textwidth]{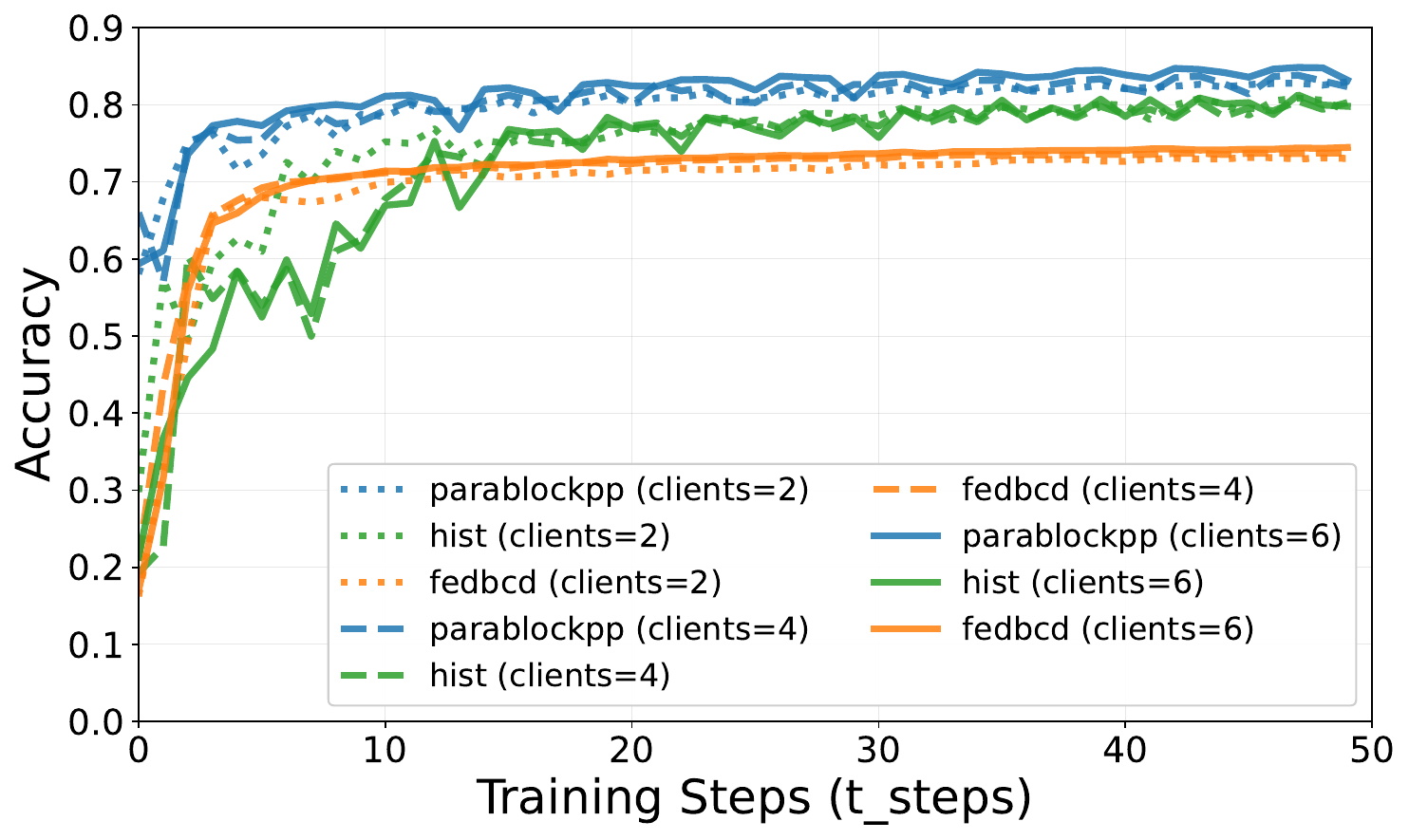}
    \vspace{10pt}
    \includegraphics[width=0.24\textwidth]{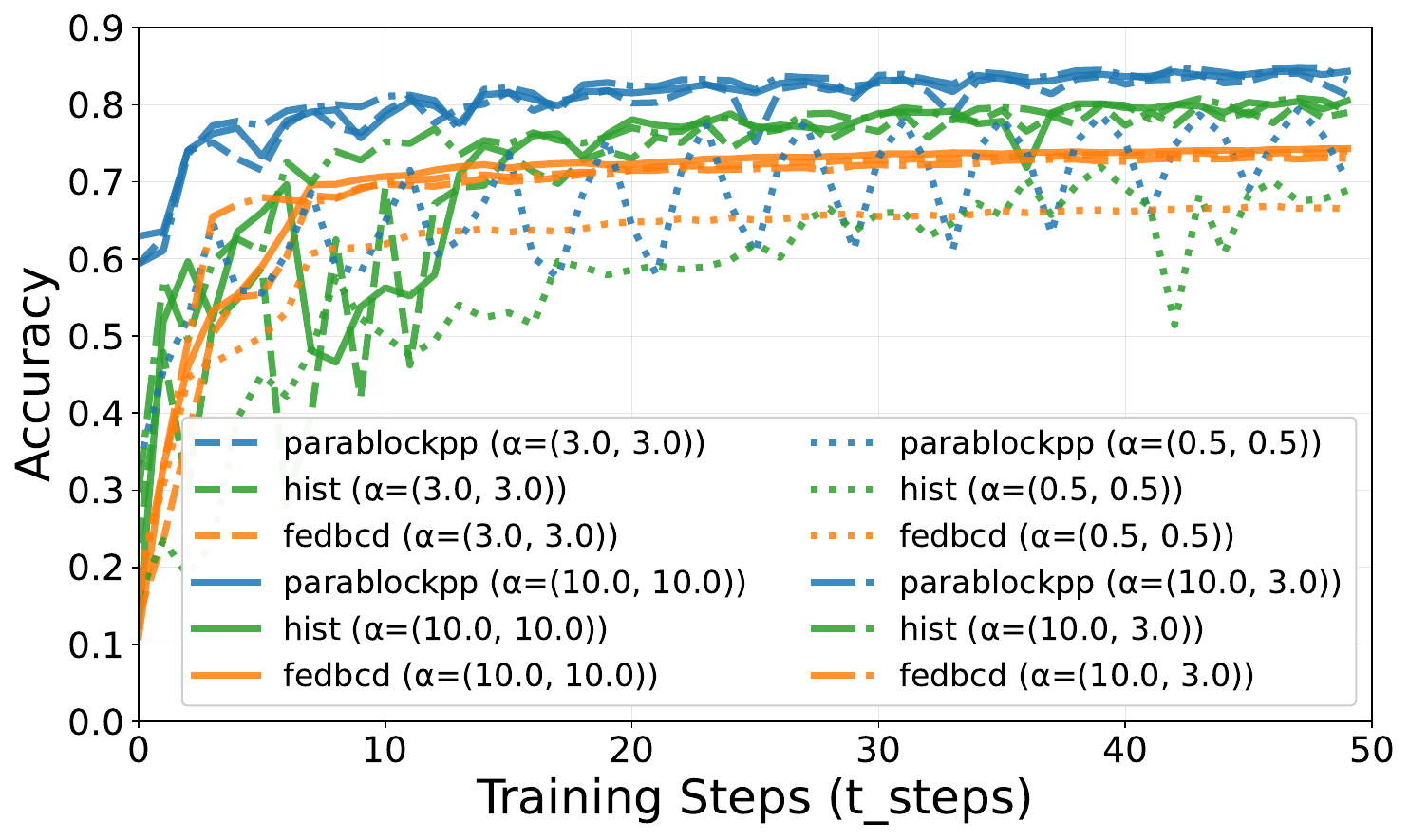}
    \includegraphics[width=0.24\textwidth]{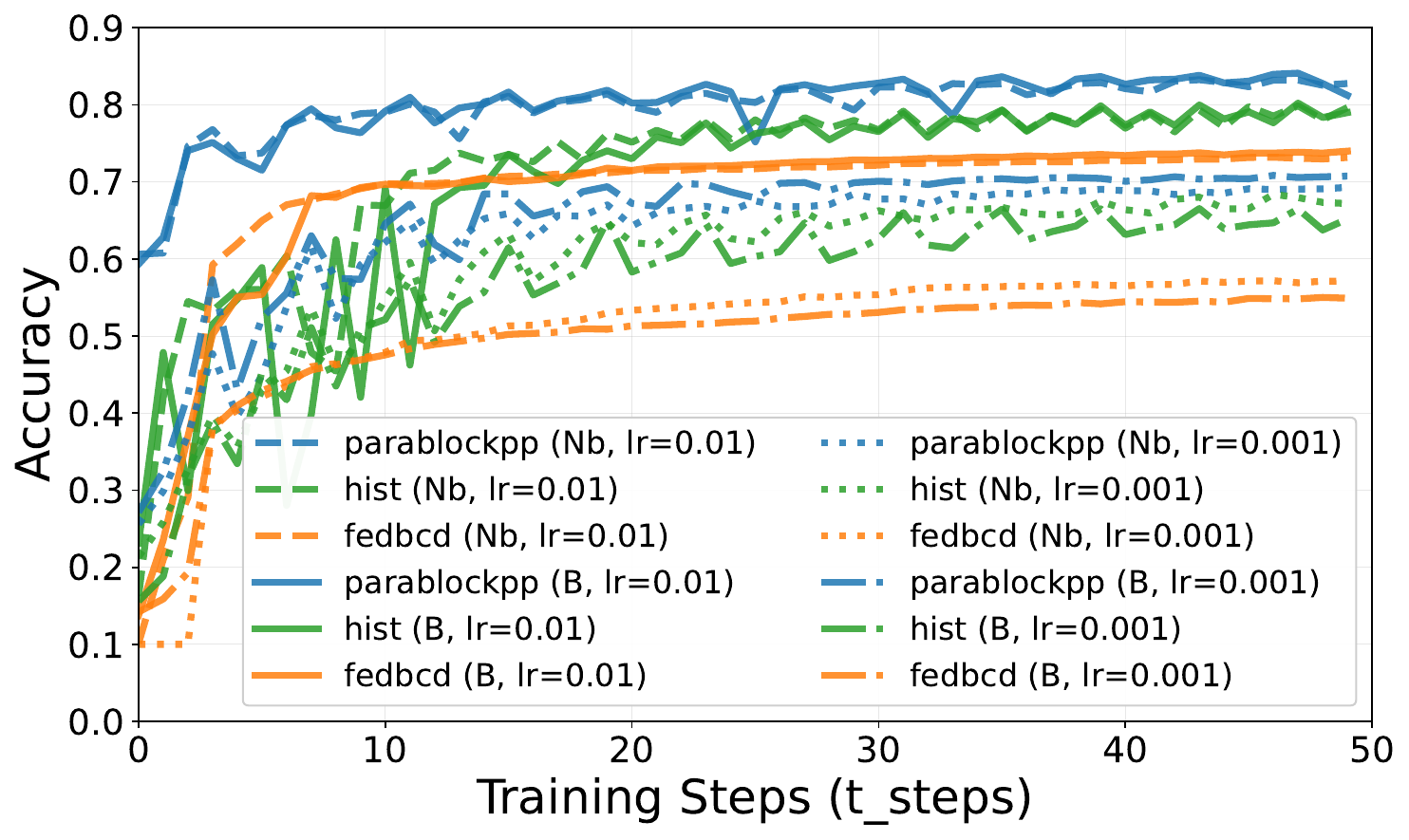}
    \vspace{-30pt}
    \caption{Comparison for ResNet-18 with the same default parameters and subplots as for LR.}
    \label{fig:resnet18}
    \vspace{-10pt}
\end{figure}

In this subsection, we benchmark $\mathrm{ParaBlock}$++ against all baselines using test accuracy versus global rounds $t$ in Figs. \ref{fig:lr}, \ref{fig:crammersingersvm}, and \ref{fig:resnet18}.

\subsubsection{Impact of $E$ and $H$}

$\mathrm{ParaBlock}$++ consistently has the highest accuracy across all configurations, with performance improving as both $E$ and $H$ increase. The highest accuracy is obtained with $E=10$ and $H=20$, which also shows the smoothest convergence. In contrast, smaller $E$ and $H$ yield lower performance with noticeable instability during training. However, FedBCD demonstrates steady convergence and minimal oscillations across all settings, whose performance improves with increasing $E$ and $H$. In contrast, HIST exhibits the most unstable training curves with significant fluctuations, particularly in early global rounds. All curves show that larger $E$ and $H$ generally result in more stable training curves and better performance. 

\subsubsection{Impact of Clients Per Cluster}

We evaluated the effect of varying the number of clients per cluster $2$, $4$, and $6$ on all models. $\mathrm{ParaBlock}$++ consistently achieves the highest accuracy, whose performance improves as the number of clients per cluster increases. The model demonstrates stable convergence across all configurations, with slightly more variability in early global rounds. FedBCD demonstrates the most stable behavior regardless of client count, maintaining consistent convergence with minimal oscillations. HIST exhibits the highest volatility across all configurations, with significant fluctuations throughout training. The results indicate that increasing clients per cluster generally improves model accuracy, with $\mathrm{ParaBlock}$++ benefiting most.

\subsubsection{Impact of Data Heterogeneity}

We evaluate the effect of different cluster-level and client-level data heterogeneity combinations. The alpha parameters control the Dirichlet distribution for data partitioning, where lower values indicate higher heterogeneity and higher values indicate more homogeneous data distribution.

$\mathrm{ParaBlock}$++ achieves the highest accuracy across all alpha configurations, with performance improving as alpha values increased. The best results were obtained with (10.0, 10.0), indicating that more homogeneous data distributions (higher alpha) benefit this algorithm. In contrast, high heterogeneity with (0.5, 0.5) significantly degrades performance, with dropping accuracy, slower convergence, and notable instability in early global rounds. FedBCD demonstrates the most stable behavior across all alpha values, maintaining consistent convergence regardless of data heterogeneity. HIST exhibits the highest volatility across all alpha values, with significant fluctuations throughout training. While higher alpha values generally improve performance, the algorithm remains highly unstable, and the volatility persists even with more homogeneous data distributions. The results demonstrate that data heterogeneity significantly impacts model performance, with more homogeneous data (higher alpha) generally leading to better accuracy. $\mathrm{ParaBlock}$++ benefits most from reduced heterogeneity, achieving substantial gains when alpha is increased. FedBCD proves highly robust to data heterogeneity, offering consistent performance across all alpha configurations. HIST remains the least reliable algorithm due to its high volatility regardless of data distribution.

\subsubsection{Impact of Partition Balance and Learning Rate}

Our experimental results reveal that $\mathrm{ParaBlock}$++ achieves the highest test accuracy across all hyperparameter configurations, while FedBCD exhibits the most stable convergence behavior. As expected, higher learning rates consistently enhance performance given a fixed number of global rounds. Notably, we observe that layer-wise partitioning has only a marginal impact on model accuracy, with block parameter ratios of 10\%, 20\%, 30\%, and 40\% for LR and CrammerSinger SVM, and 1.35\%, 4.49\%, 17.96\%, and 76.12\% for ResNet-18. Despite the pronounced imbalance in parameter ratios, performance degradation remains surprisingly limited, suggesting that the $\mathrm{ParaBlock}$++ system exhibits considerable resilience to partition skew.

\subsection{Accuracy versus Computation-Communication Time}

In this subsection, we compare $\mathrm{ParaBlock}$++ with all baselines using test accuracy versus computation–communication time $t_{cc}$ (in allocated core-seconds). Theoretically, we can obtain a direct comparison as summarized in Table \ref{tab:times}.

\begin{table}[H]
\label{tab:times}
\centering
\caption{Computation-Communication Time Per Global Round}
\begin{tabular}{ccccc}
\toprule
Model & Avg. Comp. Time & Avg. Comm. Time \\
\midrule
$\mathrm{ParaBlock}$++ & $\sum_{i\in\mathcal{I}}\bar{T}^\text{comp}_i$ & $T^\text{stsp}$ \\
HIST/FedBCD & $\sum_{i\in\mathcal{I}}\bar{T}^\text{comp}_i$ & $\max_{i\in\mathcal{I}}T^\text{dst}_i$ \\
\bottomrule
\end{tabular}
\end{table}

In Table \ref{tab:times}, $\bar{T}^\text{comp}_i$ represents the average training time of a submodel on cluster $i$, averaged across all submodels. $T^\text{stsp}$ denotes the minimum time needed to sequentially visit all cluster servers without imposing any specific order, solved as a Steiner traveling salesman problem (Steiner TSP). $T^\text{dst}_i$ indicates the time required to broadcast a submodel to all cluster servers without imposing any specific order, solved as a directed Steiner tree (DST) problem. It is evident that $T^\text{stsp}\geq\max_iT^\text{dst}_i$, as sequential transmission inherently incurs greater latency than propagation, which can potentially exploit parallelism. Nevertheless, given the hyperparameters in Table \ref{tab:simulationsetting}, simple calculations demonstrate that the accelerated convergence achieved by $\mathrm{ParaBlock}$++ sufficiently compensates for the additional communication overhead for large-sized models such as ResNet-18.

\subsection{Performance Evaluation with Various Offloading Ratios}

We propose satellite-assisted training with client data offloading, aiming to leverage satellite-side computation and decrease cluster latency. According to the formulations in Section \ref{sec:iterative}, parameters $\{\alpha_j\}_{j\in\mathcal{C}_i}$ designate the data of client $j$ offloaded to satellite $i$, having a significant impact on training latency.

\begin{figure}[htbp]
    \includegraphics[width=0.24\textwidth]{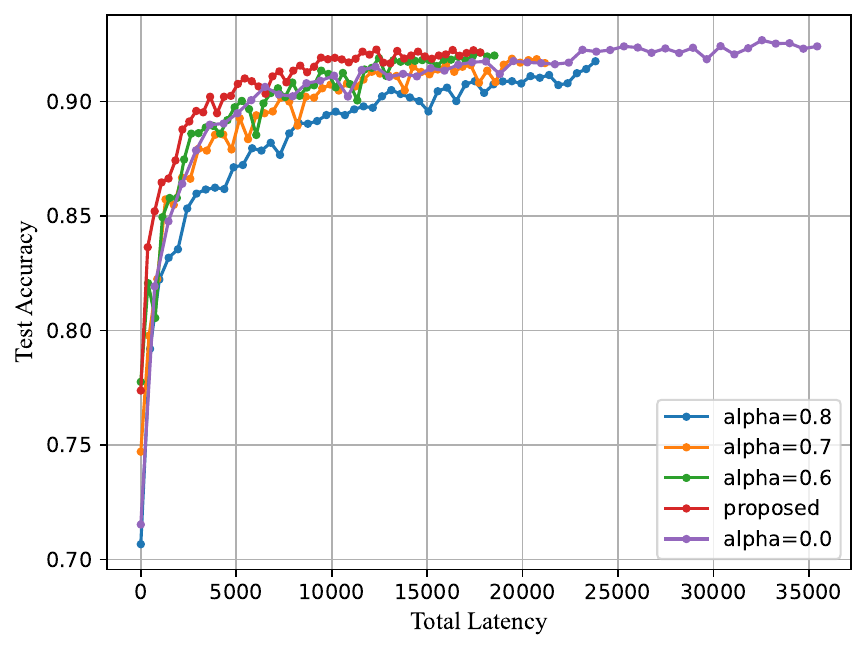}
    \includegraphics[width=0.24\textwidth]{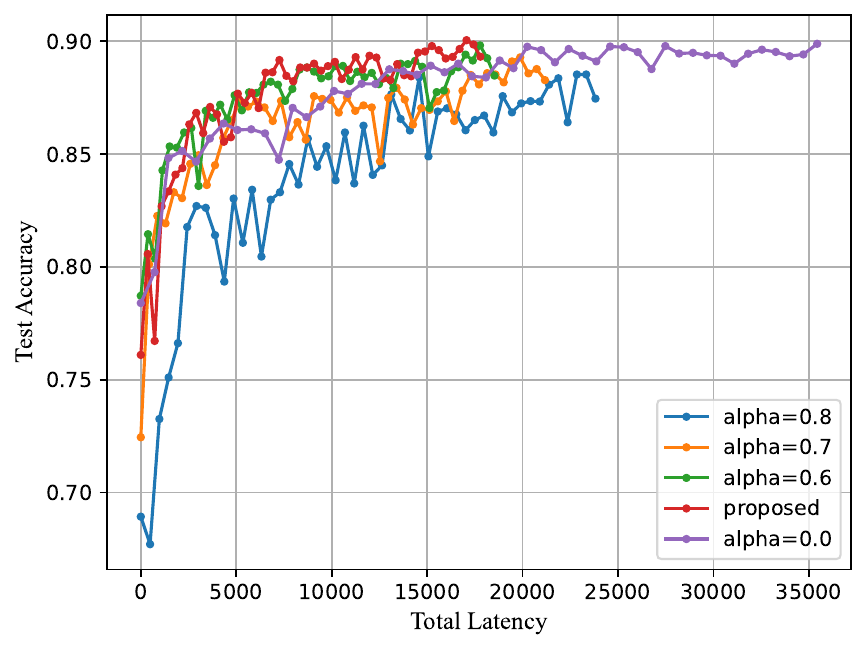}
    \vspace{-20pt}
    \caption{$\mathrm{ParaBlock}$++ for ResNet-18 with $T=50$, $E=20$, $H=20$ under i.i.d. and non-i.i.d., respectively.}
    \label{fig:alpha}
\end{figure}

Fig. \ref{fig:alpha} demonstrates and compares the overall latency in training ResNet-18 with $\mathrm{ParaBlock}$++ under i.i.d. and non-i.i.d. cluster-client data distributions, respectively. Compared with fixed ratios $\{\alpha_j\}_{j\in\mathcal{C}_i}$, solving our proposed optimization problem in Section \ref{sec:iterative} over $\{\alpha_j\}_{j\in\mathcal{C}_i}$ indeed results in minimized computation-communication latency. It is noted that different choices of ${\alpha_j}_{j\in\mathcal{C}_i}$, despite having significant impact on overall latency, only exhibit minor differences in the final performances after completing all $T$ global rounds.

\section{Conclusion and Future Work}\label{sec:conclusion}

We have presented $\mathrm{ParaBlock}$++, an HFL framework assisted by LEO satellite megaconstellations, designed to address the challenges of distributed learning in resource-constrained and geographically remote environments characteristic of future 6G networks. By leveraging LEO satellites as aggregators, participants, and high-throughput relays, and adopting asynchronous BCD-type updates, it accelerates convergence while maintaining superior performance. Our theoretical analysis establishes convergence guarantees across various objective functions, while our optimization formulations minimize training latency through efficient resource allocation and model partitioning. Experimental results validate the advantages of $\mathrm{ParaBlock}$++ in achieving rapid initial convergence and superior final accuracy on non-convex models with massive parameters such as ResNet-18, at much less computation and communication costs. These findings suggest $\mathrm{ParaBlock}$++ as a viable approach for extending FL capabilities in the 6G era, particularly for supporting AI services across remote and infrastructure-limited regions.

However, some potential issues with $\mathrm{ParaBlock}$++ remain for future research. Scalability is a core challenge and design goal in FL. Fortunately, our transmission scheme leverages training latencies to transmit all but one submodels that are time-insensitive, which are required only after the current global round. Another inherent issue with BCD-type updates is the potential delay in wall-clock times. However, many FL applications are inherently time-insensitive, which allows long latencies with BCD-type FL. For instance, FL for analyzing medical records, developing diagnostic models for chronic diseases, or conducting population-level health studies are prime candidates.


\onecolumn
\appendices

\section*{Appendix A: Technical Lemmas}


\noindent\textbf{Lemma A.1} \cite{lemmaa5}. If a nonnegative scalar sequence $\{a_k\}$ satisfies $a_{k+1} \leq(1-{A}/{k}) a_k+{B}/{k^2}$ where $A, B>0$, then it obeys $a_k \leq {c}/{k},\ \forall k \geq\lfloor A\rfloor+1$ with $c=2 B /(A-1)$ if $A>1$, and $c=\max \left\{B / A, a_1\right\}$ if $A \leq 1$.\\


\noindent\textbf{Lemma A.2} \cite{lemmaa5}. If two nonnegative scalar sequences $\{a_k\}$ and $\{b_k\}$ satisfy that $\sum_{k=1}^\infty a_k=\infty$ and $\sum_{k=1}^\infty a_kb_k <\infty$, then $\lim\inf_{k\to\infty}b_k = 0$. Furthermore, if $|b_{k+1} - b_k| \leq B \cdot a_k$ for some constant $B > 0$, then $\lim_{k\to\infty} b_k = 0$.

\section*{Appendix B: Proof of Theorem 1}

\noindent\textit{Notes on notation}. We use shorthand notations to avoid clutter where appropriate. For instance, $x_{<i}^{t+1}$ denotes $[x_{1}^{t+1},\ldots,x_{i-1}^{t+1}]$. Without loss of generality, we assume cluster $\mathcal{C}_i$ is responsible for optimizing parameter $x_i$. By convention, at all hierarchical levels, index 0 denotes the initial value immediately before the first update, i.e., immediately after parameter averaging. That is, we can define $x_{i}^{t}\!\triangleq\!x_{i,j}^{t,0,0}$ and $x_{i}^{t,e}\!\triangleq\!x_{i,j}^{t,e,0}$ for all $t\in[T]$, $e\in[E]$, $h\in[H]$, $i\in\mathcal{I}$, and $j\!\in\!\mathcal{C}_i$.\\

\noindent\textit{Proof}. Denote $(x_{1,j_1}^{t,e,h},\ldots,x_{s,j_s}^{t,e,h})$ as $x^{t,e,h}$ when we do not emphasize clients $(j_1,\ldots,j_s)$. The upper bound of $\Phi(x^{t,e,h+1}) - \Phi(x^{t,e,h})$ is given by the proximal properties as (see Eq. (21) of work \cite{bsg})
\begin{align}\label{eq:convex1}
    \Phi(x^{t,e,h+1}) - \Phi(x^{t,e,h}) &\leq
    \sum_{i=1}^s \big\langle {g}_{i,j_i}^{t,e,h} - \tilde{g}_{i,j_i}^{t,e,h}, x_i^{t,e,h+1} - x_i^{t,e,h} \big\rangle + \sum_{i=1}^s \big( \frac{L}{2} - \frac{1}{2\alpha^t} \big) \big\|x_i^{t,e,h+1} - x_i^{t,e,h}\big\|^2 \notag\\
    &\quad + \sum_{i \in \mathcal{I}_1} \Big(\big\langle \tilde{g}_{i,j_i}^{t,e,h}, x_i - x_i^{t,e,h} \big\rangle + r_i(x_i) - r_i(x_i^{t,e,h}) \Big) + \sum_{i \in \mathcal{I}_2} \big\langle \tilde{g}_{i,j_i}^{t,e,h} + \tilde{\nabla} r_i(x_i^{t,e,h}), x_i - x_i^{t,e,h} \big\rangle \notag\\
    &\quad + \frac{1}{2\alpha^t} \sum_{i=1}^s \Big(\big\|x_i - x_i^{t,e,h}\big\|^2 - \big\|x_i - x_i^{t,e,h+1}\big\|^2 \Big) + \sum_{i \in \mathcal{I}_2} 2L_{r_i} \big\|x_i^{t,e,h+1} - x_i^{t,e,h}\big\|.
\end{align}
The upper bound of $\Phi(x^{t,e,h}) - \Phi(x)$ is given by the convexity of $\Phi$ as
\begin{align}\label{eq:convex2}
    \Phi(x^{t,e,h}) - \Phi(x) &\leq
    - \sum_{i=1}^s \big\langle \nabla_{x_i} F(x^{t,e,h}), x_i - x_i^{t,e,h} \big\rangle - \sum_{i \in \mathcal{I}_1} \big(r_i(x_i) - r_i(x_i^{t,e,h})\big)
    - \sum_{i \in \mathcal{I}_2} \big\langle \tilde{\nabla} r_i(x_i^{t,e,h}), x_i - x_i^{t,e,h} \big\rangle.
\end{align}
Then, the upper bound of $\Phi(x^{t,e,h+1})-\Phi(x)$ is given by adding up Eq. \eqref{eq:convex1} and Eq. \eqref{eq:convex2} as
\begin{align}\label{eq:convex3}
    \Phi(x^{t,e,h+1})\! -\! \Phi(x) &\leq
    \sum_{i=1}^s \big\langle {g}_{i,j_i}^{t,e,h} - \tilde{g}_{i,j_i}^{t,e,h}, x_i^{t,e,h+1}\! -\! x_i^{t,e,h} \big\rangle
    + \sum_{i=1}^s \big( \frac{L}{2} - \frac{1}{2\alpha^t} \big) \big\|x_i^{t,e,h+1}\! - \!x_i^{t,e,h}\big\|^2\! +\! \sum_{i \in \mathcal{I}_2} 2L_{r_i} \big\|x_i^{t,e,h+1}\! - \!x_i^{t,e,h}\big\| \notag\\
    &\quad + \sum_{i=1}^s \big\langle \tilde{g}_{i,j_i}^{t,e,h} - \nabla_{x_i} F(x^{t,e,h}), x_i - x_i^{t,e,h} \big\rangle
    + \frac{1}{2\alpha^t} \sum_{i=1}^s \Big( \big\|x_i - x_i^{t,e,h}\big\|^2 - \big\|x_i - x_i^{t,e,h+1}\big\|^2 \Big),
\end{align}
where $\nabla_{x_i} F(x^{t,e,h})=\nabla_{x_i} F(x_{1,j_1}^{t,e,h},\ldots,x_{s,j_s}^{t,e,h})$. For all $e\in [E],h\in [H]$, similar bounding techniques in Eq. (23) of work \cite{bsg} yields
\begin{align}\label{eq:convex4}
    \mathbb{E}[\Phi(x^{t,e,h+1})] - \Phi(x) \leq
    \frac{s\big(4L_{r_i}^2+(\sigma^t)^2\big)\alpha^t}{1-L\alpha^t}
    + \sqrt{s} \big(\|x\|+\rho\big)\big(A+LM\big)\alpha^t
    + \frac{1}{2\alpha^t} \mathbb{E}\Big[ \big\|x - x^{t,e,h}\big\|^2 - \big\|x - x^{t,e,h+1}\big\|^2 \Big],
\end{align}
where the constant $M$ is slightly different from that of work \cite{bsg} and takes into account inter-cluster and intra-cluster gradient heterogeneity with the decomposition
\begin{align*}
    \nabla_{x_i}F(x^{t,e,h}) - \tilde{g}_{i,j_i}^{t,e,h}
    =\big(\nabla_{x_i}F(x^{t,e,h}) - G^{t,e,h}_{i}\big) + \big(G^{t,e,h}_{i} - {g}_i^{t,e,h}\big) + \big({g}_{i}^{t,e,h} - {g}_{i,j_i}^{t,e,h}\big) + \big({g}_{i,j_i}^{t,e,h} - \tilde{g}_{i,j_i}^{t,e,h}\big).
\end{align*}
For any $(j_1,\ldots,j_s)$, taking average of Eq. \eqref{eq:convex4} for all $h\in[H]$, we obtain
\begin{align}\label{eq:convex5}
    \mathbb{E}\Big[\frac{1}{H}\sum_{h=0}^{H-1}\Phi(x^{t,e,h+1})\Big]\! -\! \Phi(x) \leq
    \frac{s\big(4L_{r_i}^2+(\sigma^t)^2\big)\alpha^t}{1-L\alpha^t}
    + \sqrt{s} \big(\|x\|+\rho\big)\big(A+LM\big)\alpha^t
    + \frac{1}{2H\alpha^t} \mathbb{E}\Big[ \big\|x - x^{t,e,0}\big\|^2\! -\! \big\|x - x^{t,e,H}\big\|^2 \Big],
\end{align}
At time $(t,e,H)$, we compute the intra-cluster average of Eq. \eqref{eq:convex5} for all $(j_1,\ldots,j_s)$ and obtain $x^{t,e+1,0}$. By the convexity of $\Phi$ and $\|\cdot\|^2$,
\begin{align}\label{eq:convex6}
    \mathbb{E}\Big[\frac{1}{H}\big(\sum_{h=0}^{H-2}\Phi(x^{t,e,h+1})+\Phi(x^{t,e+1,0})\big)\Big] - \Phi(x) &\leq
    \frac{s\big(4L_{r_i}^2+(\sigma^t)^2\big)\alpha^t}{1-L\alpha^t}
    + \sqrt{s} \big(\|x\|+\rho\big)\big(A+LM\big)\alpha^t \notag\\
    &\quad + \frac{1}{2H\alpha^t} \mathbb{E}\Big[ \big\|x - x^{t,e,0}\big\|^2 - \big\|x - x^{t,e+1,0}\big\|^2 \Big],
\end{align}
Taking average for all $e\in[E]$ for Eq. \eqref{eq:convex6} and using the convexity of $\Phi$ and $\|\cdot\|^2$ gives
\begin{align}\label{eq:convex7}
    \mathbb{E}\Big[\frac{1}{E}\sum_{e=0}^{E-1}\frac{1}{H}\big(\sum_{h=0}^{H-2}\Phi(x^{t,e,h+1})+\Phi(x^{t,e+1,0})\big)\Big] - \Phi(x) &\leq
    \frac{s\big(4L_{r_i}^2+(\sigma^t)^2\big)\alpha^t}{1-L\alpha^t}
    + \sqrt{s} \big(\|x\|+\rho\big)\big(A+LM\big)\alpha^t \notag\\
    &\quad + \frac{1}{2EH\alpha^t} \mathbb{E}\Big[ \big\|x - x^{t,0,0}\big\|^2 - \big\|x - x^{t,E,0}\big\|^2 \Big].
\end{align}
At time $(t,E,0)$, we have completed all $E$ cluster rounds and there is no inter-cluster average in the scheme. We denote ${x}^{t+1,0,0}\triangleq{x}^{t,E,0}$ by definition, denote $\bar{x}^{t+1}$ as
\begin{align}\label{eq:convex8}
    \bar{x}^{t+1}=\frac{1}{E}\sum_{e=0}^{E-1}\frac{1}{H}\Big(\sum_{h=0}^{H-2}x^{t,e,h+1}+x^{t,e+1,0}\Big),
\end{align}
and denote $D(x)$ as
\begin{align}\label{eq:convex9}
    D(x)=\sup_t\Big\{\frac{s(4L_{r_i}^2+(\sigma^t)^2)}{1-L\alpha^t} + \sqrt{s} (\|x\|+\rho)(A+LM)\Big\}.
\end{align}
Inserting Eq. \eqref{eq:convex8} and Eq. \eqref{eq:convex9} into Eq. \eqref{eq:convex7}, the convexity of $\Phi$ gives
\begin{align}\label{eq:convex10}
    \mathbb{E}[\Phi(\bar{x}^{t+1})] - \Phi(x) &\leq D(x) + \frac{1}{2EH\alpha^t}\mathbb{E} \Big[\big\|x - x^{t,0,0}\big\|^2-\big\|x - x^{t+1,0,0}\big\|^2\Big],
\end{align}
Letting $x=x^*$ and taking Ces\`aro average for Eq. \eqref{eq:convex10} for all $t\in[T]$ with weight $\alpha^t$, the convexity of $\Phi$ gives
\begin{align}\label{eq:convex11}
    \mathbb{E}[\Phi(\tilde{x}^{T})] - \Phi(x^*) &\leq \frac{\sum_{t=0}^{T-1}(\alpha^t)^2}{\sum_{t=0}^{T-1}\alpha^t}D(x^*) + \frac{1}{2EH\sum_{t=0}^{T-1}\alpha^t} \big\|x^* - x^0\big\|^2,
\end{align}
where the Ces\`aro average $\tilde{x}^{T}=\sum_{t=1}^T\alpha^t\bar{x}^t/\sum_{t=1}^T\alpha^t$. Letting $\alpha^t=\theta/\sqrt{t+1}$,
\begin{align}\label{eq:convex12}
    \mathbb{E}[\Phi(\tilde{x}^{T})] - \Phi(x^*) &\leq \frac{\theta(\ln T+1)}{\sqrt{T+1}}D(x^*) + \frac{1}{2EH\theta\sqrt{T+1}} \big\|x^* - x^0\big\|^2.
\end{align}
This completes the proof. \hfill\qedsymbol\\

\section*{Appendix C: Proof of Theorem 2}

\noindent\textit{Proof}. For strongly convex $\Phi(x)$ with modulus $\mu$, it holds that
\begin{align}\label{eq:stronglyconvex1}
    \Phi(x^{t,e,h})-\Phi(x^*)\geq\frac{\mu}{2}\big\|x^{t,e,h}-x^*\big\|^2.
\end{align}
At time $(t,e,H)$, taking intra-cluster average of Eq. \eqref{eq:convex4} to obtain $x^{t,e+1,0}$, we have
\begin{align}\label{eq:stronglyconvex2}
    \mathbb{E}[\Phi(x^{t,e+1,0})] - \Phi(x^*) \leq
    \alpha^tD(x^*) + \frac{1}{2\alpha^t} \mathbb{E}\Big[ \big\|x^* - x^{t,e,H-1}\big\|^2 - \big\|x^* - x^{t,e+1,0}\big\|^2 \Big]
\end{align}
due to the convexity of $\Phi$ and $\|\cdot\|^2$. That is, Eq. \eqref{eq:convex4} holds true for all $t\in[T]$, $e\in[E]$, and $h\in[H]$. Then, Eq. \eqref{eq:convex4} together with Eq. \eqref{eq:stronglyconvex1} implies
\begin{align}\label{eq:stronglyconvex3}
    \mathbb{E}\Big[\big\|x^{t,e,h+1}-x^*\big\|^2\Big]
    \leq\frac{\mathbb{E}[\|x^{t,e,h+1}-x^*\|^2]}{1+\mu\alpha^t}
    +\frac{2D(x^*)\alpha^t}{1+\mu\alpha^t}
    \leq\Big(1-\frac{\mu\theta}{(t+1)(1+\mu\theta)}\Big) \mathbb{E}\Big[\big\|x^* - x^{t,e,h}\big\|^2\Big]+\frac{2D(x^*)\theta^2}{(t+1)^2}.
\end{align}
Define $k=tEH+eH+h$, then we have for $t\geq1$,
\begin{align}\label{eq:stronglyconvex4}
    \mathbb{E}\big[\|x^{k+1}-x^*\|^2\big]
    \leq&\
    \Big(1-\frac{\mu\theta}{(\lfloor k/EH\rfloor+1)(1+\mu\theta)}\Big)\mathbb{E}\big[\|x^{k}-x^*\|^2\big]
    +
    \frac{2D(x^*)\theta^2}{(\lfloor k/EH\rfloor + 1)^2} \notag\\
    \leq&\
    \Big(1-\frac{EH\mu\theta}{2k(1+\mu\theta)}\Big)\mathbb{E}\big[\|x^{k}-x^*\|^2\big]
    +
    \frac{2(EH)^2D(x^*)\theta^2}{k^2},
\end{align}
which is a recurrence to solve with Lemma A.1. For $T\geq(\lfloor A\rfloor+1)/(EH)$ with $A={EH\mu\theta}/{(2+2\mu\theta)}$, we obtain
\begin{align}\label{eq:stronglyconvex5}
    \mathbb{E}\big[\|x^T-x^*\|^2\big]\leq\frac{c}{T},\text{ where }c= \begin{cases}\displaystyle\frac{4EHD(x^*)\theta^2}{A-1}, & \text { if } A >1, \\ \max \Big\{\displaystyle\frac{2EHD(x^*)\theta^2}{A}, \displaystyle\frac{\mathbb{E}[\|x^1-x^*\|^2]}{EH}\Big\}, & \text { if } A \leq 1,\end{cases}
\end{align}
which completes the proof. \hfill\qedsymbol\\

\section*{Appendix D: Proof of Theorem 3}

\noindent\textit{Proof}. The $L$-Lipschitz continuity on $\nabla F(x)$ indicates
\begin{align}\label{eq:nonconvex1}
F(x^{t,e,h+1})-F(x^{t,e,h})
\leq&\
\sum_{i=1}^s\Big(\big\langle g^{t,e,h}_{i,j_i}, x^{t,e,h+1}_{i,j_i}-x^{t,e,h}_{i,j_i}\big\rangle
+
\frac{L}{2}\big\|x^{t,e,h+1}_{i,j_i}-x^{t,e,h}_{i,j_i}\big\|^2\Big) \notag\\
=&\
\sum_{i=1}^s\Big(-\alpha^t\big\langle g^{t,e,h}_{i,j_i}, \tilde{g}^{t,e,h}_{i,j_i}\big\rangle
+\frac{L}{2}(\alpha^t)^2\big\|\tilde{g}^{t,e,h}_{i,j_i}\big\|^2\Big) \notag\\
=&\
\sum_{i=1}^s\Big(-\big(\alpha^t-\frac{L}{2}(\alpha^t)^2\big)\big\|g^{t,e,h}_{i,j_i}\big\|^2
+\frac{L}{2}(\alpha^t)^2\big\|\delta^{t,e,h}_{i,j_i}\big\|^2
-\big(\alpha^t-L(\alpha^t)^2\big)\big\langle g^{t,e,h}_{i,j_i},\delta^{t,e,h}_{i,j_i}\big\rangle\Big) \notag\\
=&\
\sum_{i=1}^s\Big(-\big(\alpha^t-\frac{L}{2}(\alpha^t)^2\big)\big\|g^{t,e,h}_{i,j_i}\big\|^2
+\frac{L}{2}(\alpha^t)^2\big\|\delta^{t,e,h}_{i,j_i}\big\|^2
-\big(\alpha^t-L(\alpha^t)^2\big)\big\langle \nabla_{x_i}F(x^{t,e,h}),\delta^{t,e,h}_{i,j_i}\big\rangle\Big) \notag\\
&\ \qquad-\big(\alpha^t-L(\alpha^t)^2\big)\big\langle g^{t,e,h}_{i,j_i}-\nabla_{x_i}F(x^{t,e,h}),\delta^{t,e,h}_{i,j_i}\big\rangle\Big).
\end{align}
For all $e\in[E]$, $h\in[H]$, taking expectation and using the bounding techniques in Eq. (44) of work \cite{bsg}, we have
\begin{align}\label{eq:nonconvex2}
\mathbb{E}\big[F(x^{t,e,h+1})\big]-\mathbb{E}\big[F(x^{t,e,h})\big]
\leq&\
-\sum_{i=1}^s\big(\alpha^t-\frac{L}{2}(\alpha^t)^2\big)\mathbb{E}\Big[\big\|g_{i,j_i}^{t,e,h}\big\|^2\Big]+\sum_{i=1}^sM_{\rho}A\big(\alpha^t+L(\alpha^t)^2\big)\alpha^t
+\sum_{i=1}^s\mathit{O}\big((\alpha^t)^2\big),
\end{align}
where $M_\rho$ is a constant defined similar to Eq. (9) of work \cite{bsg}. Summing up for all $h\in[H]$, we obtain
\begin{align}\label{eq:nonconvex3}
\mathbb{E}\big[F(x^{t,e,H})\big]-\mathbb{E}\big[F(x^{t,e,0})\big]
=\textstyle{\sum}_{h=0}^{H-1}\big(\mathbb{E}\big[F(x^{t,e,h+1})\big]-\mathbb{E}\big[F(x^{t,e,h})\big]\big).
\end{align}
At time $(t,e,H)$, taking average of Eq. \eqref{eq:nonconvex3} over cluster $\mathcal{C}_i$ to obtain $x^{t,e+1,0}$, we have the inequality
\begin{align}\label{eq:nonconvex4}
\mathbb{E}\big[F(x^{t,e+1,0})\big]-\mathbb{E}\big[F(x^{t,e,H})\big]\leq\mathit{O}\big((\alpha^t)^2\big),
\end{align}
which follows from the $L$-Lipschitz continuity of $\nabla F(\mathbf{x})$. This implies that for all $(t,e,h+1)$, it follows that
\begin{align}\label{eq:nonconvex6}
\mathbb{E}\big[F(x^{t,e,h+1})\big]-\mathbb{E}\big[F(x^{0,0,0})\big]
{\leq}&\ 
-\sum_{(0,0,0)}^{(t,e,h)}\sum_{i=1}^s\big(\alpha^t-\frac{L}{2}(\alpha^t)^2\big)\mathbb{E}\Big[\big\|g_{i,j_i}^{t,e,h}\big\|^2\Big]+\sum_{(0,0,0)}^{(t,e,h)}\mathit{O}\big((\alpha^t)^2\big).
\end{align}
The fact that $F(x)>-\infty$ and $\sum_{t=0}^\infty(\alpha^t)^2<\infty$ indicates
\begin{align}\label{eq:nonconvex7}
\sum_{(0,0,0)}^{(t,e,h)}\alpha^t\mathbb{E}\Big[\big\|g_{i,j_i}^{t,e,h}\big\|^2\Big]<\infty,\ \forall 1\leq i\leq s.
\end{align}
Furthermore, the bounding techniques in Eq. (48) of work \cite{bsg} give
\begin{align}\label{eq:nonconvex8}
\Big|\mathbb{E}\big[\big\|g_{i,j_i}^{t,e,h+1}\big\|^2\big]-\mathbb{E}\big[\big\|g_{i,j_i}^{t,e,h}\big\|^2\big]\Big|=&\ \Big|\mathbb{E}\big[\big\|g_{i,j_i}^{t,e,h+1}+g_{i,j_i}^{t,e,h}\big\|\big\|g_{i,j_i}^{t,e,h+1}-g_{i,j_i}^{t,e,h}\big\|\big]\Big|=\mathit{O}\big(\alpha^t\big),\ \forall 1\leq i\leq s.
\end{align}
Using Lemma A.2, we have $\mathbb{E}[\|g_{i,j_i}^{t,e,h}\|^2]\to0$ as $t\to\infty$, for all $e\in[E]$, $h\in[H]$, $1\leq i\leq s$. \hfill\qedsymbol


\end{document}